%% file: main.tex
\pdfoutput=1

\documentclass[11pt]{article}

\usepackage[final]{acl}

\usepackage{times}
\usepackage{latexsym}

\usepackage[T1]{fontenc}

\usepackage[utf8]{inputenc}

\usepackage{microtype}

\usepackage{inconsolata}

\usepackage{multirow}
\usepackage{booktabs}
\usepackage{graphicx}
\usepackage{subfigure}
\usepackage{amssymb}
\usepackage{CJKutf8}
\usepackage{amsmath} 
\usepackage{hyperref}
\usepackage{enumitem}
\usepackage{color}
\graphicspath{{images/}}
\usepackage{algorithm}
\usepackage{algpseudocode}
\usepackage{tgadventor}
\usepackage{tikz}

\title{Joint Modeling of Entities and Discourse Relations for \\ Coherence Assessment}

\author{
Wei Liu \and Michael Strube \\ 
Heidelberg Institute for Theoretical Studies gGmbH  \\ 
\texttt{\{wei.liu, michael.strube\}@h-its.org}
}

\begin{document}
\maketitle
\input{1.abstract}
\input{2.introduction}
\input{3.related_work}
\input{4.method}

\input{5.experiments}

\input{6.analysis}

\input{7.conclusion}
\input{8.limitations}

\section*{Acknowledgements}
The authors would like to thank the three anonymous reviewers for their comments. This work has been funded by the Klaus Tschira Foundation, Heidelberg, Germany.

\bibliography{custom}

\clearpage
\appendix
\input{9.appendix}

\end{document}

%% file: 1.abstract.tex
\begin{abstract}
In linguistics, coherence can be achieved by different means, such as by maintaining reference to the same set of entities across sentences and by establishing discourse relations between them. However, most existing work on coherence modeling focuses exclusively on either entity features or discourse relation features, with little attention given to combining the two. In this study, we explore two methods for jointly modeling entities and discourse relations for coherence assessment. Experiments on three benchmark datasets show that integrating both types of features significantly enhances the performance of coherence models, highlighting the benefits of modeling both simultaneously for coherence evaluation.

\end{abstract}

%% file: 2.introduction.tex
\section{Introduction}
Coherence is a property of well-written texts that makes them easier to read and understand than a sequence of randomly strung sentences~\citep{lapatacoh}. Its modeling benefits many downstream NLP tasks, such as machine translation~\citep{sia-duh-2023-context}, topic modeling~\citep{li-etal-2023-diversity}, text generation~\citep{guan2023generating}, and dialog generation~\citep{mendonca-etal-2024-ecoh}.

In linguistics, text coherence can be achieved in several ways, with two of the most widely studied being entity-based and discourse relation-based coherence~\citep{coherencecondition,speechlanguage}. Entity-based coherence focuses on how entities are introduced and maintained throughout a text~\citep{Prince_Taxonomy81,grosz-etal-1995-centering}. In contrast, discourse relation-based coherence considers the logical or rhetorical relationships between sentences~\citep{kehler08,rohde-etal-2018-discourse}. These perspectives have inspired distinct modeling approaches: entity-based methods~\citep{barzilay-lapata-2008-modeling,guinaudeau-strube-2013-graph,tien-nguyen-joty-2017-neural,jeon-strube-2022-entity} typically model local coherence by tracking entity transitions, while discourse-based methods~\citep{lin-etal-2011-automatically,feng-etal-2014-impact,wang-etal-2019-using,wu-etal-2023-multi-task} evaluate coherence based on parsed discourse relations.

\begin{figure}[t]
\centering\includegraphics[scale=0.45,trim=0 0 0 0]{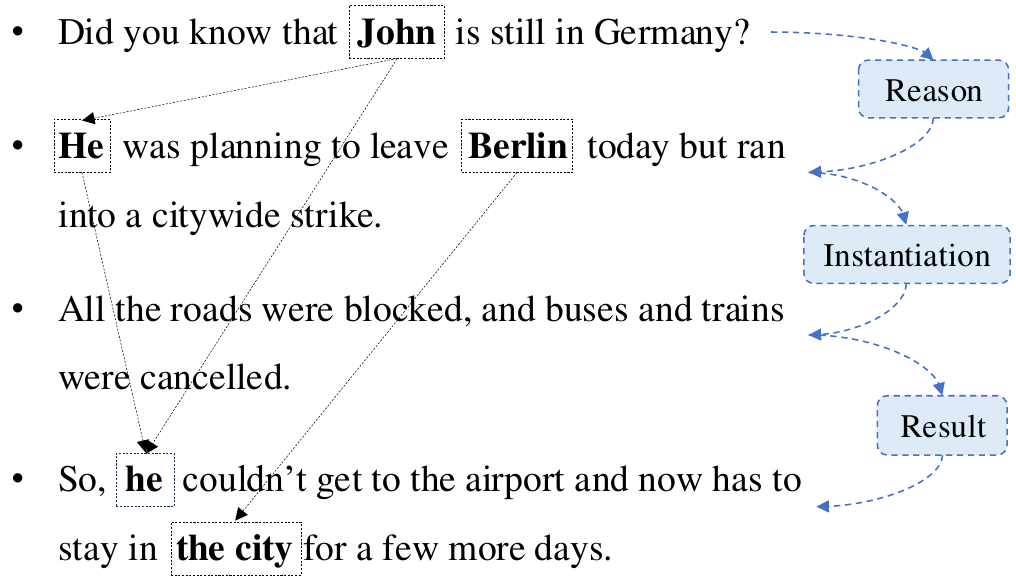}
\setlength{\abovecaptionskip}{0pt}
\setlength{\belowcaptionskip}{0pt}
\caption{An example of a coherent text, whose coher\-ence should be explained using both entities and discourse relations. We bold the interlinked entities in the text and show the discourse relations between sentences.}
\label{fig:example}
\end{figure}

While these approaches have proven effective individually, real-world texts often require a more integrated view. In practice, entity and discourse relation cues frequently coexist and interact in complex ways. To illustrate this, we present an example in Figure \ref{fig:example}, which contains four sentences and is considered highly coherent. Establishing the coherence using entities is not straightforward in this case, as there are no overlapping entities between the second and third sentences. Instead, we must use a more complex linguistic phenomenon, namely bridging~\citep{clark-1975-bridging,hou-etal-2018-unrestricted}, to link ``city'' (in ``citywide'') and ``road''. Meanwhile, the connection between these sentences is more readily explained by a discourse relation (e.g., Instantiation), as the third sentence elaborates on the strike mentioned earlier. However, relying solely on discourse relations also has limitations, as it can compromise the smooth tracking of the protagonist if the referents are unclear. For example, if the final sentence were changed to ``So, Maria couldn't get to the airport...'' the discourse relation\- might still hold, but the referent switch (i.e., John $\to$ Maria) would disrupt the overall coherence. This underscores the need to jointly consider both entity continuity and discourse structure. Despite their complementary nature, few studies have empirically investigated whether combining these two perspectives leads to more effective coherence assessment.

To address this gap, we propose two approaches for jointly modeling entities and discourse relations in coherence evaluation. The first approach identifies the entities in a document and the discourse relations between sentences, then organizes them, along with the sentences, in a flat structure. We introduce a fusion Transformer that jointly models these elements to assess coherence. The second approach avoids dedicated fusion modules by incorporating entity and discourse relation information directly into prompts, allowing large language models (LLMs) to leverage them during inference.

We evaluate\footnote{\url{https://github.com/liuwei1206/EntyRelCoh}} our methods on three benchmarks: two for assessing discourse coherence and one for automatic essay scoring. Our models significantly outperform strong baselines, demonstrating the benefits of joint modeling. Further analysis reveals that integrating both entities and discourse relations enables better learning of coherence patterns, which help to mitigate the effects of imbalanced data distributions in datasets and improve models' generalization across domains.


%% file: 3.related_work.tex
\section{Related Work}

Our work is related to existing approaches that enhance coherence modeling using entities, discourse relations, or Transformer-based models.

\noindent \textbf{Entity-based}. The most well-known entity-based model is the Entity Grid, proposed by \citet{barzilay-lapata-2008-modeling}, which constructs a two-dimensional matrix to capture the transitions of entities between adjacent sentences. This model has been improved by various subsequent efforts, such as incorporating semantically related entities~\citep{filippova-strube-2007-extending} and integrating entity-specific features~\citep{elsner-charniak-2011-extending}. Another prominent entity-centered approach is the Entity Graph, proposed by \citet{guinaudeau-strube-2013-graph}, which measures textual coherence by evaluating the extent to which sentences are connected to each other via shared discourse entities. Building on similar ideas, \citet{mesgar-strube-2015-graph,mesgar-strube-2016-lexical} model coherence using the local connectivity structure of sentences. With the rise of deep learning, neural networks have also been applied to capture entity-based coherence patterns. For example, \citet{tien-nguyen-joty-2017-neural} and \citet{joty-etal-2018-coherence} extend the entity grid using convolutional neural networks. \citet{jeon-strube-2020-centering} introduce a structure-aware model to approximate Centering Theory, which is further refined by \citet{jeon-strube-2022-entity} through the use of more linguistically grounded units, such as noun phrases and proper names.

\noindent \textbf{Discourse Relation-based}. Compared to entity-based models, fewer studies have employed discourse relations for coherence assessment, largely due to the limited performance of early discourse parsers. One of the earliest works in this area is by \citet{lin-etal-2011-automatically}, who use discourse relations as features for evaluating coherence. Specifically, they adopt an approach similar to the entity grid, constructing a two-dimensional matrix where rows represent sentences and columns represent entities, and each cell ($s_i$, $e_j$) contains the set of discourse roles of the entity $e_j$ that appears in the sentence $s_i$. \citet{feng-etal-2014-impact} extend this approach by replacing shallow discourse relations with deeper ones derived from an RST~\cite{RST} parser. However, \citet{mesgar-strube-2015-graph} criticize these methods as conceptually flawed, arguing that treating discourse relations as features of entities contradicts their linguistic function, which is to link sentences or elementary discourse units (EDUs). More recently, \citet{wu-etal-2023-multi-task} propose a multi-task framework that jointly identifies discourse relations between sentences and evaluates the overall coherence of a text.

Unlike these two lines of work focusing solely on entities or discourse relations, we aim to combine both for more effective coherence modeling.

\noindent \textbf{Transformer-based}. Our work is also related to recent studies that use Transformer models for coherence assessment. \citet{abhishek2021transformer} demonstrate that RoBERTa significantly outperforms earlier embedding-based models, with performance further improving under a multi-task training setup incorporating NLI tasks. \citet{laban-etal-2021-transformer} use Transformer models to tackle the shuffle test task, achieving near-perfect accuracy (97.88\%). To probe the capabilities of language models in coherence prediction, \citet{beyer-etal-2021-incoherence} design targeted test suites addressing diverse aspects of discourse and dialogue coherence. Building on these directions, \citet{zhao-etal-2023-discoscore} propose DiscoScore, a BERT-based metric inspired by Centering Theory, which models coherence from multiple discourse perspectives and shows a high correlation with human judgments across coherence and factual consistency. More recently, large language models have also been applied to coherence evaluation. \citet{naismith-etal-2023-automated} show that GPT-4 can produce coherence ratings comparable to those of human annotators, accompanied by well-reasoned explanations. Similarly, \citet{mansour-etal-2024-large} assess ChatGPT and LLaMA on essay scoring tasks, finding that, with appropriate prompting, both models achieve strong performance even in one-shot settings.

\begin{figure}[t]
\centering\includegraphics[scale=0.48,trim=0 0 0 0]{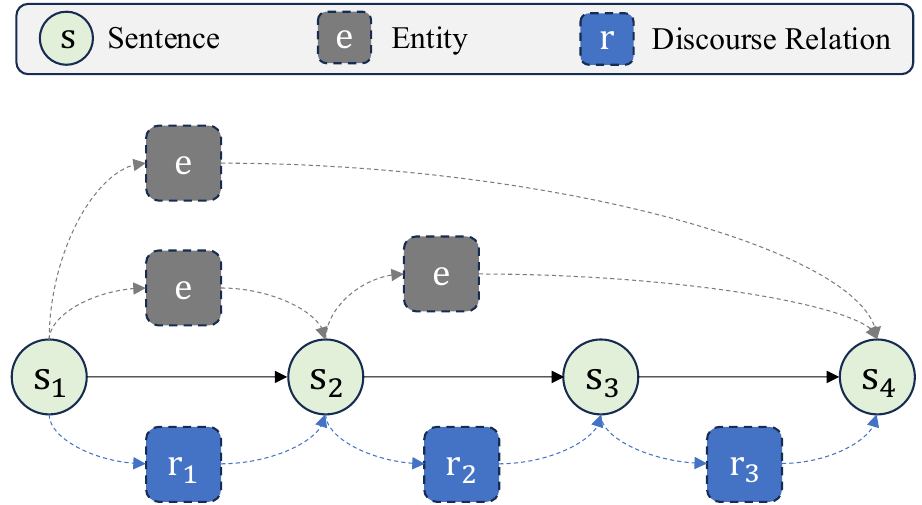}
\setlength{\abovecaptionskip}{12pt}
\setlength{\belowcaptionskip}{0pt}
\caption{Sentences (in Figure \ref{fig:example}) linked by entities and discourse relations.}
\label{fig:sententyrel}
\end{figure}

%% file: 4.method.tex
\section{Method}
In this section, we introduce how to identify entities and discourse relations in a document, followed by two methods that use the identified entities and discourse relations to evaluate coherence.

Given a document, we use Stanza~\citep{qi-etal-2020-stanza} to identify all nouns and co-references, and to segment the text into sentences. We focus on nouns rather than entities because previous studies have shown that using nouns leads to better performance in coherence modeling~\citep{elsner-charniak-2011-extending,tien-nguyen-joty-2017-neural}. For discourse relations, we follow prior work\citep{lin-etal-2011-automatically} that adopts the Penn Discourse Treebank (PDTB) framework~\citep{pdtb2}. Specifically, we use the discourse parser {\tt discopy}, developed by \citet{knaebel-2021-discopy}, to extract relations between adjacent sentences, with a few modifications. First, we use PDTB 3.0~\citep{pdtb3} instead of PDTB 2.0~\citep{pdtb2}, as the former includes more relation types and is an improved version of the latter. Second, for implicit discourse relation classification, we use the model proposed by \citet{liu-strube-2023-annotation}, which achieves state-of-the-art performance. We provide more details about the parser in Appendix \ref{app:pdtb_parser}.

After identifying nouns, coreference relations, and discourse relations, we link two sentences if: (1) they share the same nouns or there is a coreference link between mentions in the sentences, or (2) they are connected by a discourse relation. In the first case, we add an edge labeled ``entity'' between the sentences, while in the second case, we add an edge labeled with the specific type of discourse relation. Figure \ref{fig:sententyrel} shows how the sentences in Figure \ref{fig:example} are linked through the identified entities and discourse relations, forming a graph structure. 


However, since the Transformer is designed for sequence modeling~\citep{transformer}, it doesn't naturally handle graph-structured input. One possible solution is to use Graph Neural Networks (GNNs), but standard GNNs are permutation-invariant and cannot capture order information~\citep{SurveyGNN}, which is crucial for coherence modeling~\citep{lapata-2003-probabilistic}. Below we introduce two approaches to address these issues.

\subsection{Method I: Fusion}
\begin{figure}[t]
\centering\includegraphics[scale=0.43,trim=0 0 0 0]{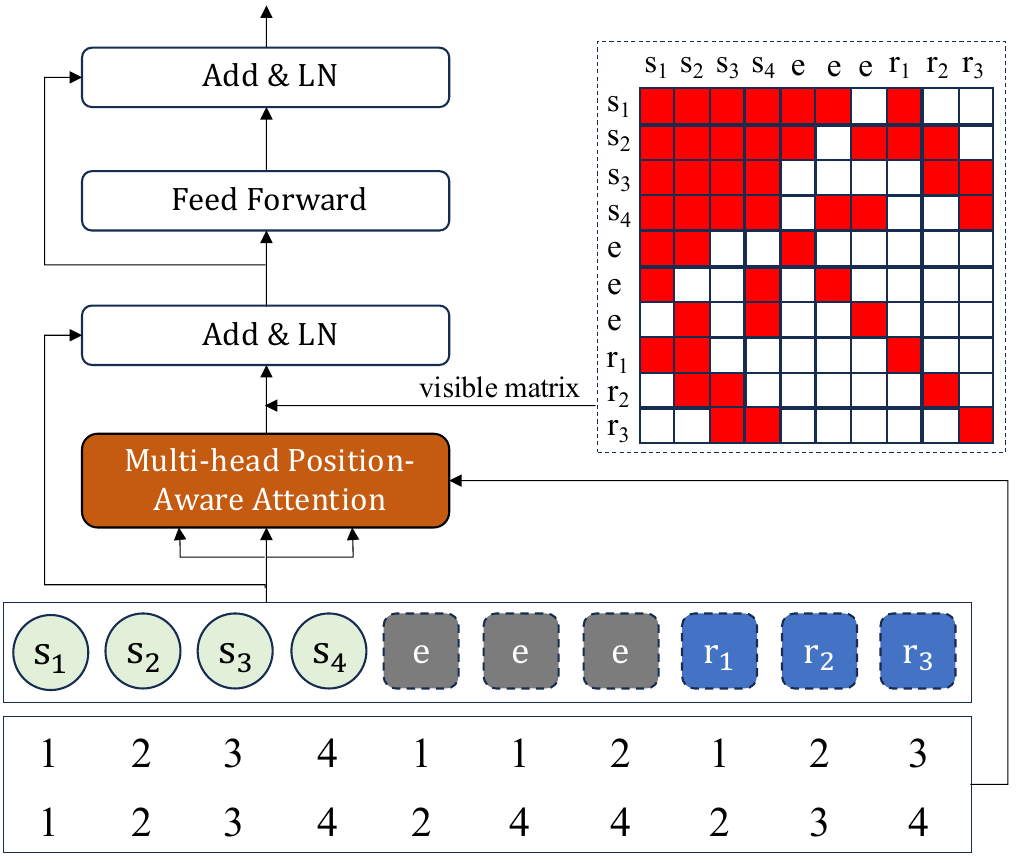}
\setlength{\abovecaptionskip}{12pt}
\setlength{\belowcaptionskip}{0pt}
\caption{The sentences, entities, and discourse relations in Figure \ref{fig:sententyrel} are organized into a flat structure, in which each element is assigned a two-dimensional position, indicating its start and end position in the original sentence sequence. This flat input is then processed by a fusion Transformer.}
\label{fig:fusion}
\end{figure}

In this approach, we introduce a flat structure to organize sentences, entities, and discourse relations, and design a fusion transformer to jointly model these elements. Figure \ref{fig:fusion} shows an overview.

In the flat structure, sentences, entities, and discourse relations are concatenated into a sequence. Each element in this sequence is assigned a two-dimensional position (see the bottom part in Figure \ref{fig:fusion}), indicating its \textbf{start} and \textbf{end} positions within the original sentence sequence. Take $\rm s_1$ and $\rm r_1$ for an example, their positions are $\rm (1, 1)$ and $\rm (1, 2)$, respectively, which means that $\rm s_1$ is the first sentence\- in the text and $\rm r_1$ links the first and second sentences. This flat structure preserves sentence order as well as the connections among sentences, entities, and discourse relations. Its sequential format also makes it well-suited for Transformer models.


\begin{figure*}[t]
\centering\includegraphics[scale=0.425,trim=0 0 0 0]{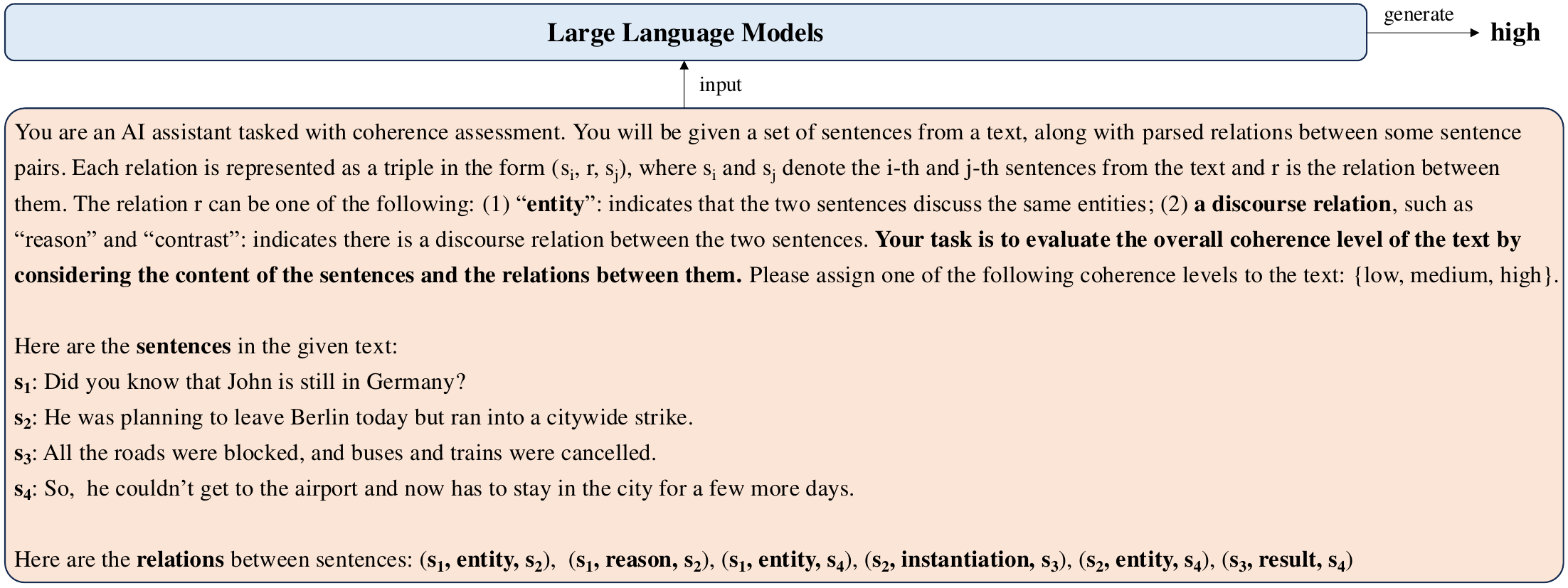}
\setlength{\abovecaptionskip}{0pt}
\setlength{\belowcaptionskip}{0pt}
\caption{Illustration of our second approach. We use natural language to describe the relationships between sentences, entities, and discourse relations in Figure \ref{fig:sententyrel}, presenting the graph structure in a concise and intuitive way. We then instruct LLMs to consider these elements for coherence assessment.}
\label{fig:prompt}
\end{figure*}

To handle this flat structure, we propose a fusion Transformer that enhances the vanilla Transformer with a novel position-aware attention mechanism and a visible matrix. Specifically, we first use a text encoder, such as RoBERTa or LLama, to obtain the representations of sentences, entities, and discourse relations. Then, we input all the elements along with their two-dimensional positions into the position-aware attention. The position-aware attention between the $i$-th and the $j$-th elements in the sequence is defined as:
\begin{equation}
    \label{eq-attn}
    \mathbf{A}_{ij}=\mathbf{q}_i\mathbf{k}_j^T + \mathbf{q}_i\mathbf{r}_{i-j}^T + \mathbf{u}\mathbf{k}_j^T + \mathbf{v}\mathbf{r}_{i-j}^T 
\end{equation}
where $\mathbf{q}_i, \mathbf{k}_j, \mathbf{r}_{i-j}=\mathbf{e}_i\mathbf{W}_q, \mathbf{e}_j\mathbf{W}_k, \mathbf{pe}_{i-j}\mathbf{W}_r$, $\mathbf{e}_i$ means the representation of the $i$-th element, $\mathbf{pe}_{i-j}$ denotes the relative position embedding between the $i$-th and the $j$-th elements, and $\mathbf{W}_q$, $\mathbf{W}_k$, $\mathbf{W}_r$, $\mathbf{u}$, $\mathbf{v}$ are trainable parameters. The first and third terms in Eq. \ref{eq-attn} are content-based addressing\-, where the former calculates weight between query and key, and the latter governs a global content bias~\citep{dai-etal-2019-transformer}. The second and last terms compute weight with relative positional information, which can be used to guide the attention between relevant elements. Since each element in the flat structure has a 2D position, we can calculate four types of relative distances between the $i$-th and $j$-th elements: (i) $\rm \mathbf{s}tart_i - \mathbf{s}tart_j$; (ii) $\rm \mathbf{s}tart_i - \mathbf{e}nd_j$; (iii) $\rm \mathbf{e}nd_i - \mathbf{s}tart_j$; (iv) $\rm \mathbf{e}nd_i - \mathbf{e}nd_j$. The final relative position embedding between the $i$-th and $j$-th elements, i.e., $\mathbf{pe}_{i-j}$, is defined as a non-linear transformation over the four relative distances:
\begin{equation}
    \mathbf{pe}_{i-j}=(\mathbf{p}_{s_{i}-s_{j}}\otimes\mathbf{p}_{s_{i}-e_{j}}\otimes\mathbf{p}_{e_{i}-e_{j}}\otimes\mathbf{p}_{e_{i}-e_{j}})\mathbf{W}_p
\end{equation}
The position embedding $\mathbf{p}$ is initialized as in Transformer, where $\mathbf{p}_{pos}^{2k}=\sin\left(pos/10000^{2k/d_{model}}\right)$ and $\mathbf{p}_{pos}^{2k+1}=\cos\left(pos/10000^{2k/d_{model}}\right)$. 


To prevent sentences from attending to irrelevant entities and discourse relations, we further introduce a visible matrix $\mathbf{M}$ to guide the attention:
\begin{equation}
    \mathbf{M}_{ij}= \begin{cases} 0, & \text {if $\rm C_1 \;|\; C_2 \;|\; C_3 \;|\; C_4$} \\ -\infty, & \text{otherwise} \end{cases}
\end{equation}

\noindent where $\rm C_1$ is $i=j$ (i.e., self-connection), $\rm C_2$ is that both $i$-th and $j$-th elements are sentences (text content), $\rm C_3$ is that one element is a sentence and the other is an entity, and the sentence links to the entity (entity patterns), and $\rm C_4$ is defined as nodes $i$ and $j$ is one sentence and one relation, and the relation works on the sentence (discourse relation patterns). We apply the visible matrix to the attention calculation:
\begin{equation}
    \mathbf{A}^* = {\rm Softmax}(\mathbf{A}+\mathbf{M})
\end{equation}
Then layer normalizations and a feed-forward network (as shown in Figure \ref{fig:fusion}) are applied to produce the text representation. Finally, we input the representation into a softmax classifier, and use the cross-entropy loss for training.

\subsection{Method II: Prompt}
While the first approach can model coherence using entity and discourse relation information, it relies on an additional fusion module and cannot fully leverage the generative capabilities of Large Language Models (i.e., it merely treats LLMs as a feature extractor). Inspired by \citet{ye-etal-2024-language}, we explore a second approach that uses natural language to describe the connections among sentences, entities, and discourse relations, and then prompts LLMs to take these information into account for coherence assessment. Figure \ref{fig:prompt} illustrates this approach using the example from Figure \ref{fig:example} and its corresponding connection graph from Figure \ref{fig:sententyrel}.


Given a graph composed of sentences, entities, discourse relations, and their connections, we traverse all sentence nodes in the order they appear in the text, from left to right. Sentences are added to the prompt and labeled with their position (e.g., $\rm s_1$, $\rm s_2$, etc., see Figure \ref{fig:prompt}). For each sentence node, we perform a depth-first search to find all two-hop neighboring nodes that are bridged by an entity or a discourse relation. This allows us to break down the graph into a list of triples, where each triple ($\rm s_i$, $\rm r_{ij}$, $\rm s_j$) includes two sentences, $\rm s_i$ and $\rm s_j$, along with the relation $\rm r_{ij}$ between them. We only retain triples where $\rm i < j$, following the natural left-to-right reading order of humans, as suggested by \citet{liu-etal-2023-modeling}. For example, the graph in Figure \ref{fig:sententyrel} is broken down into the following triples: ($\rm s_1$, entity, $\rm s_2$), ($\rm s_1$, reason, $\rm s_2$), ($\rm s_1$, entity, $\rm s_4$), ($\rm s_2$, instantiation, $\rm s_3$), ($\rm s_2$, entity, $\rm s_4$), ($\rm s_3$, result, $\rm s_4$). These triples are expressed in natural language format, making them easy for LLMs to process. More importantly, they retain all the connection information between sentences, entities, and discourse relations. Finally, we include the list of triples in the prompt and instruct the LLMs to assess coherence by considering both the content of the sentences and the patterns of entities and discourse relations between them (see Figure \ref{fig:prompt}).


\begin{table*}[t]
\centering
\scalebox{0.885}{
\renewcommand{\arraystretch}{1.2}
\begin{tabular}{lll|ccccc|c}
\hline
\multicolumn{3}{l|}{\multirow{2}{*}{Model}}                                                                        & \multicolumn{5}{c|}{\textbf{GCDC}}               & \multicolumn{1}{l}{\multirow{2}{*}{\textbf{CoheSentia}}} \\ \cline{4-8}
\multicolumn{3}{l|}{}                                                                                              & Clinton\hspace{0.6em} & Enron\hspace{0.8em} & Yahoo\hspace{0.8em} & Yelp\hspace{0.8em}  & Avg   & \multicolumn{1}{l}{}                            \\ \hline
\multicolumn{3}{l|}{\citet{jeon-strube-2022-entity}}                                                                        & 64.20\textsubscript{0.4} & 55.30\textsubscript{0.3} & 58.40\textsubscript{0.2} & 57.30\textsubscript{0.2} & 58.90 & -                                               \\
\multicolumn{3}{l|}{\citet{liu-etal-2023-modeling}}                                                                             & 66.20\textsubscript{0.8} & 57.00\textsubscript{0.8} & 63.65\textsubscript{0.7} & 58.05\textsubscript{1.2} & 61.23 & -                                               \\ \hline \hline
\multicolumn{1}{l|}{\multirow{8}{*}{Fusion}} & \multicolumn{1}{l|}{\multirow{4}{*}{RoBERTa}}          & TextOnly   & 64.55\textsubscript{0.7} & 57.50\textsubscript{0.9} & 60.05\textsubscript{0.4} & 58.20\textsubscript{0.8} & 60.10 & 60.64\textsubscript{1.5}                                           \\
\multicolumn{1}{l|}{}                        & \multicolumn{1}{l|}{}                                  & TextEnty   & 66.20\textsubscript{0.8}   & 58.80\textsubscript{1.1} & 63.15\textsubscript{0.9} & 59.20\textsubscript{1.1} & 61.83 & 63.13\textsubscript{2.0}                                           \\
\multicolumn{1}{l|}{}                        & \multicolumn{1}{l|}{}                                  & TextRel    & 66.45\textsubscript{0.9}   & 59.70\textsubscript{1.0} & 63.35\textsubscript{1.1} & 60.40\textsubscript{1.3} & 62.48 & 63.74\textsubscript{1.8}                                           \\
\multicolumn{1}{l|}{}                        & \multicolumn{1}{l|}{}                                  & Our Method I & \textbf{67.60}\textsubscript{0.5} & 60.50\textsubscript{0.3} & 63.75\textsubscript{0.5} & 61.10\textsubscript{0.4} & \textbf{63.24} & 66.24\textsubscript{1.6}                                          \\ \cline{2-9} 
\multicolumn{1}{l|}{}                        & \multicolumn{1}{l|}{\multirow{4}{*}{Llama}}            & TextOnly   & 63.55\textsubscript{0.5}   & 56.65\textsubscript{0.8} & 59.45\textsubscript{0.8} & 57.45\textsubscript{1.0} & 59.27 & 63.13\textsubscript{1.2}                                           \\
\multicolumn{1}{l|}{}                        & \multicolumn{1}{l|}{}                                  & TextEnty   & 64.80\textsubscript{0.8}   & 58.10\textsubscript{0.4} & 62.10\textsubscript{0.5} & 57.90\textsubscript{0.8} & 60.73  & 65.80\textsubscript{1.5}                                           \\
\multicolumn{1}{l|}{}                        & \multicolumn{1}{l|}{}                                  & TextRel    & 65.10\textsubscript{0.7}   & 58.75\textsubscript{0.4} & 62.85\textsubscript{0.3} & 59.35\textsubscript{0.5} & 61.51 & 66.65\textsubscript{1.6}                                           \\
\multicolumn{1}{l|}{}                        & \multicolumn{1}{l|}{}                                  & Our Method I & 67.25\textsubscript{0.4} & 60.10\textsubscript{0.3} & \textbf{64.10}\textsubscript{0.5} & \textbf{61.30}\textsubscript{0.5} & 63.18 & \textbf{69.12}\textsubscript{1.5}                                           \\ \hline \hline
\multicolumn{1}{l|}{\multirow{8}{*}{Prompt}} & \multicolumn{1}{l|}{\multirow{4}{*}{Llama zero-shot}}  & TextOnly   & 54.50\hspace{1.0em}   & 38.00\hspace{1.0em} & 34.00\hspace{1.0em} & 40.50\hspace{1.0em} & 40.88 & 50.10\hspace{1.0em}                                           \\
\multicolumn{1}{l|}{}                        & \multicolumn{1}{l|}{}                                  & TextEnty   & 55.00\hspace{1.0em}   & 39.00\hspace{1.0em} & 41.50\hspace{1.0em} & 44.50\hspace{1.0em} & 45.00 & 51.35\hspace{1.0em}                                           \\
\multicolumn{1}{l|}{}                        & \multicolumn{1}{l|}{}                                  & TextRel    & 57.50\hspace{1.0em}   & 41.00\hspace{1.0em} & 42.00\hspace{1.0em} & 45.50\hspace{1.0em} & 46.50 & 52.17\hspace{1.0em}                                           \\
\multicolumn{1}{l|}{}                        & \multicolumn{1}{l|}{}                                  & Our Method II & 56.50\hspace{1.0em}   & 41.00\hspace{1.0em} & 42.00\hspace{1.0em} & 48.00\hspace{1.0em} & 46.88 & 53.83\hspace{1.0em}                                           \\ \cline{2-9} 
\multicolumn{1}{l|}{}                        & \multicolumn{1}{l|}{\multirow{4}{*}{Llama fine-tuned}} & TextOnly   & 63.55\textsubscript{0.8}   & 56.80\textsubscript{0.9} & 60.05\textsubscript{1.0} & 55.45\textsubscript{1.2} & 58.96 & 64.95\textsubscript{1.4}                                          \\
\multicolumn{1}{l|}{}                        & \multicolumn{1}{l|}{}                                  & TextEnty   & 65.00\textsubscript{1.2}   & 57.60\textsubscript{0.5} & 60.45\textsubscript{1.0} & 56.30\textsubscript{0.9} & 59.84 & 65.38\textsubscript{1.5}                                          \\
\multicolumn{1}{l|}{}                        & \multicolumn{1}{l|}{}                                  & TextRel    & 64.55\textsubscript{0.7}   & 59.10\textsubscript{0.5} & 61.10\textsubscript{0.7} & 57.25\textsubscript{0.5} & 60.50 & 66.42\textsubscript{1.4}                                           \\
\multicolumn{1}{l|}{}                        & \multicolumn{1}{l|}{}                                  & Our Method II & 65.15\textsubscript{0.6}   & \textbf{60.55}\textsubscript{1.2} & 62.05\textsubscript{1.2} & 57.55\textsubscript{0.5} & 61.33 & 67.28\textsubscript{1.1}                                          \\ \hline
\end{tabular}}
\setlength{\abovecaptionskip}{12pt}
\setlength{\belowcaptionskip}{0pt}
\caption{Mean accuracy results (with std) on GCDC and CoheSentia.}
\label{table:gcdc_cohe}
\end{table*}

%% file: 5.experiments.tex
\section{Experiments}
\label{sec:experiment}
\noindent \textbf{Datasets}. We conduct experiments on three widely used corpora in coherence modeling: GCDC~\citep{lai-tetreault-2018-discourse}, CoheSentia~\citep{maimon-tsarfaty-2023-cohesentia}, and TOEFL~\citep{toefl11}. GCDC is a corpus designed for evaluating discourse coherence, containing texts from four distinct domains: \textbf{Yahoo} online forum posts, \textbf{Enron} emails, emails from Hillary \textbf{Clinton}’s office, and \textbf{Yelp} business reviews. Each text in the dataset is rated by experts on a scale of 1 to 3, indicating low, medium, and high levels of coherence. CoheSentia is another dataset used to assess discourse coherence. Unlike GCDC, which consists of real-world texts, CoheSentia contains stories generated by GPT-3 and is annotated by humans with coherence scores ranging from 1 to 5. However, the score distribution is highly imbalanced,\footnote{Over 50\% of the data is labeled with a score of 5.} which makes it difficult for models to converge during training~\citep{maimon-tsarfaty-2023-cohesentia}. To address this, we group scores 1 and 2 as low coherence, scores 3 and 4 as medium coherence, and score 5 as high coherence. The TOEFL dataset was originally created for automated essay scoring but has since been widely used to evaluate coherence models~\citep{burstein-etal-2010-using,jeon-strube-2020-centering}. It includes essays written in response to eight prompts (P1 to P8) along with score levels (low/medium/high) for each essay.

\noindent \textbf{Implementation Details}. We implement our models using the PyTorch library. For Method I, we experiment with two widely used text encoders~\citep{abhishek2021transformer,parmar-etal-2024-towards}: the pre-trained language model $\rm RoBERTa_{base}$~\citep{roberta} and the large language model $\rm Llama\text{-}3.1\text{-}8B\text{-}Instruction$~\citep{grattafiori2024llama3herdmodels}.\footnote{We use the 8B LLaMA model instead of the 70B due to memory limitations that prevent fine-tuning larger models. However, our resources do support zero-shot experiments with the 70B model. To maintain consistency across settings, we use the 8B model throughout the main text, but include zero-shot results for the 70B model in the Appendix \ref{app:zero-shot}.\label{why8b}} Training is performed using the AdamW optimizer with an initial learning rate of 1e-3, a batch size of 32, and a maximum of 20 epochs.

For Method II, which is specifically designed for large language models (LLMs), we evaluate it using $\rm Llama\text{-}3.1\text{-}8B\text{-}Instruction$.\textsuperscript{\ref{why8b}} The evaluation is conducted under two settings: \textbf{zero-shot} and \textbf{fine-tuned}. In the \textbf{zero-shot} setting, the model is not trained beforehand; instead, it is directly prompted to generate labels. This setup tests whether incorporating entity and discourse relation features can help with coherence evaluation in cold-start scenarios. In the \textbf{fine-tuned} setting, we fine-tune the Llama model using LoRA for 3 epochs, with a learning rate of 5e-5 and a batch size of 2. This setup evaluates whether instruction-tuning the LLM to consider entities and discourse relations can enhance its performance.


To account for training variability, we perform 10-fold cross-validation on the GCDC training dataset~\citep{lai-tetreault-2018-discourse}, 5-fold cross-validation on the CoheSentia corpus, and 5-fold cross-validation on the dataset for each prompt in the TOEFL corpus~\citep{taghipour-ng-2016-neural}. Following prior work, we use standard accuracy (Acc, \%) as our primary evaluation metric.\footnote{We also report the results of Macro-F1 in Appendix \ref{app:f1}.}


\begin{table*}[t]
\centering
\Large
\scalebox{0.585}{
\renewcommand{\arraystretch}{1.2}
\begin{tabular}{lll|ccccccccc}
\hline
\multicolumn{3}{l|}{Model}                                                                                            & P1\hspace{0.9em}    & P2\hspace{0.9em}    & P3\hspace{0.9em}    & P4\hspace{0.9em}    & P5\hspace{0.9em}    & P6\hspace{0.9em}    & P7\hspace{0.9em}    & P8\hspace{0.9em}    & Avg   \\ \hline
\multicolumn{3}{l|}{\citet{jeon-strube-2022-entity}}                                                                           & 78.38\hspace{0.9em} & 75.70\hspace{0.9em} & 76.58\hspace{0.9em} & 76.56\hspace{0.9em} & 79.10\hspace{0.9em} & 76.41\hspace{0.9em} & 75.03\hspace{0.9em} & 74.54\hspace{0.9em} & 76.54  \\
\multicolumn{3}{l|}{\citet{liu-etal-2023-modeling}}                                                                                & 75.79\textsubscript{1.1} & 76.25\textsubscript{1.1} & 74.14\textsubscript{1.2} & 75.81\textsubscript{0.7} & 77.01\textsubscript{0.9} & 77.08\textsubscript{1.1} & 73.55\textsubscript{0.8} & 72.91\textsubscript{0.7} & 75.34 \\ \hline \hline
\multicolumn{1}{l|}{\multirow{8}{*}{Fusion}} & \multicolumn{1}{l|}{\multirow{4}{*}{RoBERTa}}          & TextOnly      & 76.36\textsubscript{0.9} & 75.10\textsubscript{1.0} & 75.29\textsubscript{0.5} & 75.33\textsubscript{1.5} & 75.90\textsubscript{1.0} & 75.61\textsubscript{1.9} & 73.76\textsubscript{0.9} & 73.34\textsubscript{1.1} & 75.08 \\
\multicolumn{1}{l|}{}                        & \multicolumn{1}{l|}{}                                  & TextEnty      & 79.05\textsubscript{1.4} & 77.15\textsubscript{1.2} & 77.73\textsubscript{0.8} & 76.98\textsubscript{1.3} & 77.64\textsubscript{1.6} & 78.32\textsubscript{1.5} & 76.49\textsubscript{1.3} & 75.79\textsubscript{1.0} & 77.39\\
\multicolumn{1}{l|}{}                        & \multicolumn{1}{l|}{}                                  & TextRel       & 78.94\textsubscript{0.8} & 77.41\textsubscript{0.7} & 77.80\textsubscript{0.8} & 77.55\textsubscript{0.8} & 78.49\textsubscript{0.9} & 78.33\textsubscript{1.5} & 77.08\textsubscript{1.2} & 76.25\textsubscript{0.5} & 77.73 \\
\multicolumn{1}{l|}{}                        & \multicolumn{1}{l|}{}                                  & Our Method I  & 79.92\textsubscript{0.8} & \textbf{78.46}\textsubscript{0.9} & \textbf{78.68}\textsubscript{0.9} & \textbf{78.25}\textsubscript{1.2} & 79.23\textsubscript{1.1} & 79.42\textsubscript{1.27} & 78.21\textsubscript{0.9} & 77.13\textsubscript{1.1} & \textbf{78.66} \\ \cline{2-12} 
\multicolumn{1}{l|}{}                        & \multicolumn{1}{l|}{\multirow{4}{*}{Llama}}            & TextOnly      & 75.17\textsubscript{0.8} & 73.88\textsubscript{1.3} & 73.63\textsubscript{1.6} & 73.67\textsubscript{1.4} & 75.89\textsubscript{1.0} & 75.10\textsubscript{0.9} & 73.67\textsubscript{1.4} & 72.87\textsubscript{1.5} & 74.24 \\
\multicolumn{1}{l|}{}                        & \multicolumn{1}{l|}{}                                  & TextEnty      & 77.03\textsubscript{0.8} & 75.59\textsubscript{1.4} & 75.14\textsubscript{1.5} & 75.20\textsubscript{1.5} & 77.07\textsubscript{0.9} & 77.12\textsubscript{0.8} & 75.48\textsubscript{0.6} & 74.17\textsubscript{1.4} & 75.85 \\
\multicolumn{1}{l|}{}                        & \multicolumn{1}{l|}{}                                  & TextRel       & 76.35\textsubscript{0.9} & 76.40\textsubscript{0.7} & 75.98\textsubscript{0.5} & 75.40\textsubscript{1.2} & 76.64\textsubscript{1.7} & 76.65\textsubscript{1.6} & 75.18\textsubscript{1.1} & 75.16\textsubscript{1.3} & 75.97 \\
\multicolumn{1}{l|}{}                        & \multicolumn{1}{l|}{}                                  & Our Method I  & 78.24\textsubscript{1.7} & 78.11\textsubscript{1.9} & 77.01\textsubscript{1.1} & 76.59\textsubscript{1.1} & 79.23\textsubscript{1.3} & \textbf{79.47}\textsubscript{1.6} & 77.32\textsubscript{1.1} & 76.50\textsubscript{1.8} & 77.81 \\ \hline \hline
\multicolumn{1}{l|}{\multirow{8}{*}{Prompt}} & \multicolumn{1}{l|}{\multirow{4}{*}{Llama zero-shot}}  & TextOnly      & 51.39\hspace{0.9em} & 55.19\hspace{0.9em} & 52.72\hspace{0.9em} & 50.63\hspace{0.9em} & 54.37\hspace{0.9em} & 50.62\hspace{0.9em} & 46.92\hspace{0.9em} & 49.44\hspace{0.9em} & 51.41 \\
\multicolumn{1}{l|}{}                        & \multicolumn{1}{l|}{}                                  & TextEnty      & 56.85\hspace{0.9em} & 53.78\hspace{0.9em} & 54.48\hspace{0.9em} & 54.00\hspace{0.9em} & 53.83\hspace{0.9em} & 57.15\hspace{0.9em} & 55.89\hspace{0.9em} & 54.64\hspace{0.9em} & 55.08 \\
\multicolumn{1}{l|}{}                        & \multicolumn{1}{l|}{}                                  & TextRel       & 58.51\hspace{0.9em} & 56.45\hspace{0.9em} & 54.73\hspace{0.9em} & 55.59\hspace{0.9em} & 56.43\hspace{0.9em} & 57.19\hspace{0.9em} & 57.41\hspace{0.9em} & 53.72\hspace{0.9em} & 56.25 \\
\multicolumn{1}{l|}{}                        & \multicolumn{1}{l|}{}                                  & Our Method II & 59.90\hspace{0.9em} & 57.75\hspace{0.9em} & 56.73\hspace{0.9em} & 56.13\hspace{0.9em} & 57.28\hspace{0.9em} & 58.02\hspace{0.9em} & 58.19\hspace{0.9em} & 55.91\hspace{0.9em} & 57.49 \\ \cline{2-12} 
\multicolumn{1}{l|}{}                        & \multicolumn{1}{l|}{\multirow{4}{*}{Llama fine-tuned}} & TextOnly      & 79.03\textsubscript{1.1} & 76.76\textsubscript{1.4} & 76.24\textsubscript{1.5} & 77.52\textsubscript{1.4} & 79.49\textsubscript{1.4} & 76.02\textsubscript{1.4} & 76.69\textsubscript{1.1} & 75.28\textsubscript{0.9} & 77.13 \\
\multicolumn{1}{l|}{}                        & \multicolumn{1}{l|}{}                                  & TextEnty      & \textbf{80.13}\textsubscript{1.2} & 76.63\textsubscript{1.2} & 75.64\textsubscript{1.3} & 77.73\textsubscript{1.0} & 79.55\textsubscript{1.5} & 76.57\textsubscript{1.6} & \textbf{78.95}\textsubscript{1.4} & 76.41\textsubscript{1.3} & 77.70 \\
\multicolumn{1}{l|}{}                        & \multicolumn{1}{l|}{}                                  & TextRel       & 79.35\textsubscript{1.5} & 77.15\textsubscript{1.6} & 77.16\textsubscript{1.4} & 76.61\textsubscript{1.2} & 80.15\textsubscript{1.1} & 75.41\textsubscript{1.5} & 78.29\textsubscript{1.3} & 76.89\textsubscript{1.4} & 77.63 \\
\multicolumn{1}{l|}{}                        & \multicolumn{1}{l|}{}                                  & Our Method II & 80.02\textsubscript{1.6} & 77.92\textsubscript{1.5} & 77.58\textsubscript{1.2} & 78.13\textsubscript{1.3} & \textbf{81.13}\textsubscript{1.5} & 77.29\textsubscript{1.3} & 77.88\textsubscript{1.0} & \textbf{77.18}\textsubscript{1.5} & 78.39 \\ \hline
\end{tabular}}
\setlength{\abovecaptionskip}{10pt}
\setlength{\belowcaptionskip}{-4pt}
\caption{Mean accuracy results (with std) on TOEFL dataset.}
\label{table:toefl}
\end{table*}

\noindent \textbf{Baselines}. To validate the importance of modeling entities and discourse relations simultaneously, we compare it with the following baselines:
\begin{itemize}
    \item \textbf{TextOnly}. This baseline relies solely on textual information for coherence modeling. In Method I, this involves using a text encoder to obtain sentence representations, a sentence-level transformer to capture coherence patterns, and a softmax classifier for prediction. In Method II, it prompts LLMs to evaluate coherence based only on the text.

    \item \textbf{TextEnty}. This is an ablated version of our approach in which the discourse relation elements are removed from the sentence-entity-discourse relation graph.

    \item \textbf{TextRel}. This is another ablated version of our method, where we remove the entity elements from the graph.
\end{itemize}
\vspace{-2pt}
\noindent Further, we compare our approaches against previous state-of-the-art models on each corpus. For more details on the datasets, implementation, and baselines, please refer to Appendix \ref{app:setting}.


\subsection{Overall Results}
\noindent \textbf{GCDC / CoheSentia}. Table \ref{table:gcdc_cohe}
shows the results on GCDC and CoheSentia datasets, where the ``Fusion'' block shows the results relying on an extra fusion module to integrate entity and discourse relation features, while the ``Prompt'' block presents the results using natural languages to incorporate entity and discourse relation patterns into the input prompt of LLMs.

For the Fusion style, we show the results based on RoBERTa and LLama. Regardless of whether RoBERTa or Llama is used as the text encoder, TextEnty and TextRel consistently outperform the TextOnly baseline on GCDC and CoheSentia. This suggests that incorporating entity or discourse relation features enhances coherence assessment, which is in line with the findings of previous entity-based~\citep{jeon-strube-2022-entity} and discourse relation-based studies~\citep{wu-etal-2023-multi-task}. The improvement of TextRel over TextOnly is greater than that of TextEnty over TextOnly. This is because, in both GCDC and CoheSentia, discourse relations are more commonly used to connect sentences than entity cues. For instance, discourse relations like cause and concession are frequently employed in CoheSentia to make stories more compact and engaging~\citep{chaturvedi-etal-2017-story}. Our Method I significantly outperforms both the TextEnty\- and TextRel baselines, showing a 1\% to 2\% improvement on GCDC and approximately a 3\% gain on CoheSentia. These results highlight the value of jointly modeling entity and discourse relation features for effective coherence assessment.

\begin{figure*}[t]
\centering\includegraphics[scale=0.40,trim=0 0 0 0]{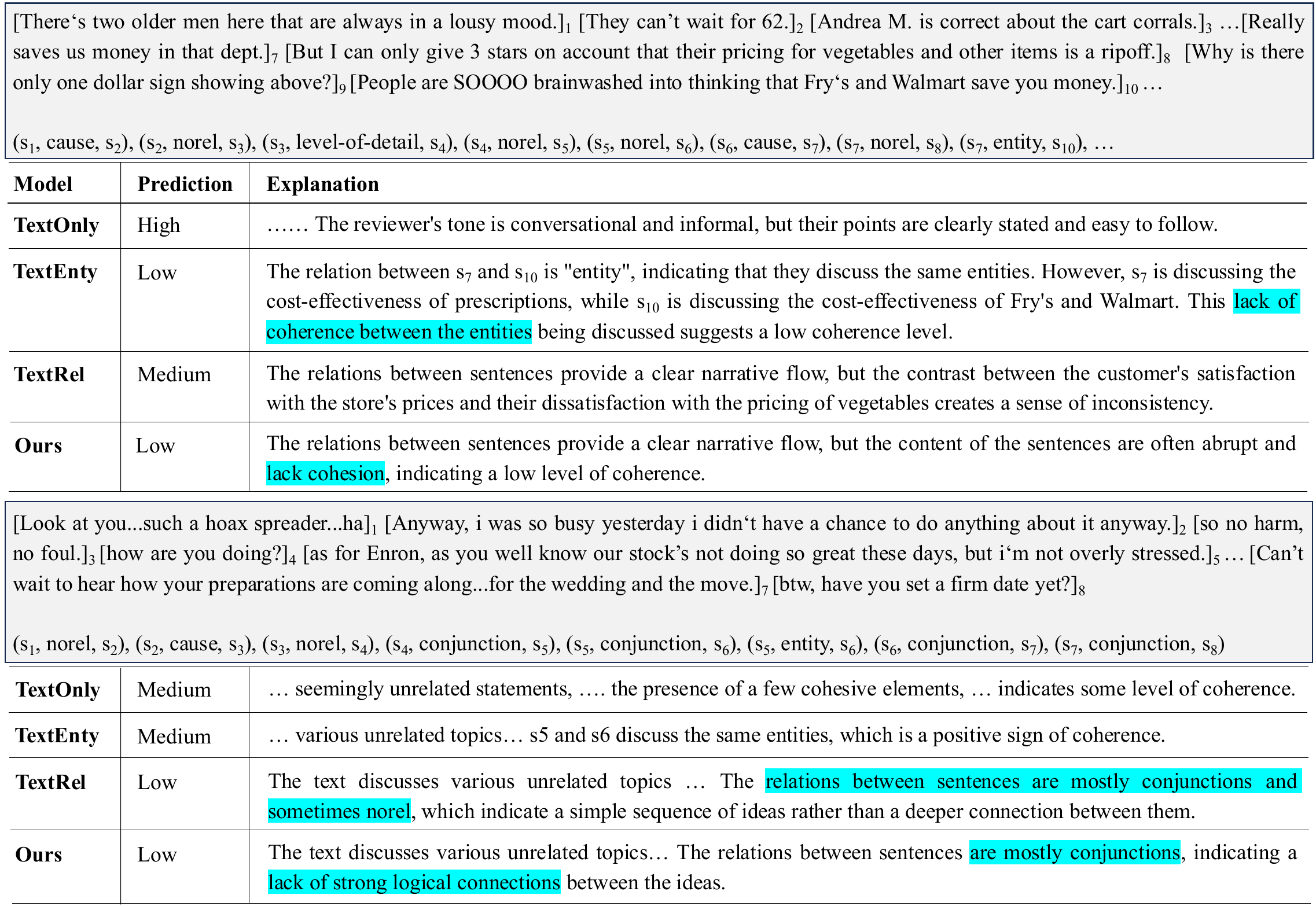}
\setlength{\abovecaptionskip}{6pt}
\setlength{\belowcaptionskip}{0pt}
\caption{Two examples (truncated) showing how entities and discourse relations aid coherence assessment. Both texts are labeled as low coherence. We use a zero-shot prompt setting, and the "explanation" refers to Llama’s brief justification for its prediction.}
\label{fig:case_study}
\end{figure*}

For the Prompt style, we present the results of Llama in both zero-shot and fine-tuned settings. In the zero-shot setting, incorporating entity and discourse relation information enhances Llama's performance in coherence assessment. On GCDC, TextEnty~and~TextRel outperform the TextOnly baseline by more than 4\% to 5\%. In contrast, the improvement on CoheSentia is more modest, with gains of about 1\% to 2\%. Combining these features further boosts performance, leading to improvements of over 6 points on GCDC and 3.5\% on CoheSentia, compared to the TextOnly baseline. These results suggest that prior knowledge of entity- and discourse relation-based coherence can be effectively leveraged for coherence assessment in cold-start scenarios. When fine-tuning LLaMA with LoRA, the performance improvements of \mbox{TextEnty}, TextRel, and EntyRel over \mbox{TextOnly} still exists, but the gains are smaller compared to the zero-shot setting. We speculate that this is because fine-tuning allows the model to somewhat \mbox{implicitly} capture coherence-relevant signals, such as entity transition and discourse relations~\citep{xiao-etal-2021-predicting}, so the explicit incorporation of them leads to limited improvement.

\noindent \textbf{TOEFL}. Results on TOEFL are shown in Table \ref{table:toefl}. Similar to the findings on GCDC and \mbox{CoheSentia}, both entity and discourse relation patterns contribute positively to the task in the fusion setting. Specifically, TextEnty~and~TextRel outperform the TextOnly baseline by 2\% to 3\% when using RoBERTa or Llama as the text encoder. Combining entity and discourse relation features further enhances performance. Our Method I using RoBERTa as the text encoder achieves an average accuracy of 78.66\%, significantly outperforming the previous state-of-the-art model~\citep{jeon-strube-2022-entity}. We observe similar results in the prompt setting: in the zero-shot scenario, Method II achieves an accuracy of 6.08\% higher than the TextOnly baseline, and 1.26\% higher in the fine-tuned setting.

%% file: 6.analysis.tex
\section{Analysis}
\label{sec:analysis}
To understand how jointly modeling entities and discourse relations contributes to coherence assessment, we analyze the accuracy of different models across each coherence label. Table \ref{table:per_label} presents the models' performance on the TOEFL P5 dataset in both the Fusion setting (with Llama as the text encoder) and the fine-tuned Prompt setting. TextOnly exhibits a strong bias, achieving high accuracy on ``medium'' and ``high'' coherence labels but significantly lower accuracy on the ``low'' label. We attribute this to the imbalanced label distribution in the TOEFL P5 dataset, where over 90\% of samples are annotated as ``medium'' or ``high'' coherence. TextEnty and TextRel help mitigate this bias by incorporating entity and discourse relation information, respectively. For example, in the Fusion setting, they improve accuracy on low-coherence data by 6.57\% and 7.69\%. Our Methods I and II go further by jointly modeling entities and discourse relations, resulting in the smallest performance gap across all three coherence levels. These results suggest that incorporating entities and discourse relations helps the model learn more effective coherence patterns and improves its robustness to imbalanced data distributions.

\begin{table}[t]
\centering
\Large
\scalebox{0.56}{
\renewcommand{\arraystretch}{1.2}
\begin{tabular}{l|l|ccc|r}
\hline
                        &               & Low   & Medium & High & Range \\ \hline
\multirow{4}{*}{\begin{tabular}[c]{@{}l@{}}Fusion \\ (Llama)\end{tabular}} & TextOnly      & 66.67 & 78.99  & 77.88 & 12.32 \\
                        & TextEnty      & 73.24 & 80.44  & 76.79 & 7.20 \\
                        & TextRel       & 74.36 & 80.45  & 78.41 & 6.09 \\
                        & Our Method I  & 81.16 & 81.99  & 77.19 & 4.80 \\ \hline
\multirow{4}{*}{\begin{tabular}[c]{@{}l@{}}Prompt \\ (fine-tuned)\end{tabular}} & TextOnly      & 68.22 & 83.29  & 82.93 & 15.07 \\
                        & TextEnty      & 71.70 & 85.23  & 85.49 & 13.79 \\
                        & TextRel       & 70.59 & 84.09  & 84.05 & 13.50 \\
                        & Our Method II & 73.47 & 85.39  & 84.71 & 11.92 \\ \hline
\end{tabular}}
\setlength{\abovecaptionskip}{6pt}
\setlength{\belowcaptionskip}{-14pt}
\caption{Accuracy results for each coherence label on TOEFL P5. Range indicates the difference between the highest and lowest values.}
\label{table:per_label}
\end{table}

To better understand how entities and discourse relations influence model behavior, we present two case studies in Figure \ref{fig:case_study}. The two examples are from GCDC corpus and annotated as low coherence. In both cases, we use a zero-shot prompt setting, asking Llama to evaluate the coherence level of a given text and provide a brief explanation for its assessment (see Appendix \ref{app:explanation} for details). As shown in the first example, without entity and discourse relation information (i.e., TextOnly), Llama evaluates the text as having high coherence. TextRel\- identifies some inconsistencies but still fails to classify it as medium coherence. In contrast, TextEnty and Our Method II correctly assess the text as having low coherence, due to the lack of cohesion, specifically, missing entity-based signals. In the second example, all models recognize that the sentences in the text cover various unrelated topics. However, TextOnly and TextEnty are slightly influenced by the presence of cohesive elements, leading them to predict the text as medium coherence. In contrast, TextRel and Our Method II correctly and confidently classify it as low coherence, due to the lack of logical connections between the sentences. These two cases effectively illustrate the importance of modeling both entity and discourse relation patterns for accurate coherence assessment.

\begin{table}[t]
\centering
\Large
\scalebox{0.51}{
\renewcommand{\arraystretch}{1.2}
\begin{tabular}{l|l|c|c}
\hline
&              & Enron $\to$ Others & TOEFL P1 $\to$ Others \\ \hline
\multirow{4}{*}{\begin{tabular}[c]{@{}l@{}}Fusion \\ (Llama)\end{tabular}}     & TextOnly     & 47.48\hspace{3.2em}                      & 68.79\hspace{3.2em}                         \\
& TextEnty     & 50.62 (+3.14)              & 72.02 (+3.23)                 \\
& TextRel      & 50.98 (+3.55)              & 72.87 (+4.08)                 \\
& Our Method I & 53.82 (+6.34)              & 74.40 (+5.61)                 \\ \hline
\multirow{4}{*}{\begin{tabular}[c]{@{}l@{}}Prompt\\ (fine-tuned)\end{tabular}} & TextOnly     & 52.50\hspace{3.2em}                      & 76.72\hspace{3.2em}                         \\
& TextEnty     & 53.67 (+1.17)              & 78.42 (+1.70)                 \\
& TextRel      &54.75 (+2.25)              & 78.15 (+1.43)                 \\
& Our Method II & 56.00 (+3.50)              & 78.60 (+1.88)                 \\ \hline
\end{tabular}}
\setlength{\abovecaptionskip}{6pt}
\setlength{\belowcaptionskip}{-17pt}
\caption{Accuracy of models in a cross-domain setting.}
\label{table-transfer}
\end{table}

To assess whether our models have truly learned more robust coherence patterns, we further evaluate their transferability in cross-domain settings. Specifically, we train TextOnly, TextEnty, TextRel, and Our Method in both Fusion and Prompt settings on the Enron subset of GCDC (or Prompt 5 of TOEFL) and test their performance on other subsets of GCDC (or other TOEFL prompts). Table \ref{table-transfer} presents the results. Both TextEnty and TextRel consistently outperform the TextOnly baseline in cross-domain settings, indicating that entity and discourse relation patterns are effective domain-agnostic features for coherence assessment. Moreover, our methods achieve the best performance across all cross-domain experiments, demonstrating the effectiveness of jointly modeling entities and discourse relations.

%% file: 7.conclusion.tex
\section{Conclusions}
This paper explores whether combining entity and discourse relation information improves coherence modeling. We propose two novel methods that jointly model entities and discourse relations for coherence assessment. Experiments on three benchmark datasets show that our approaches consistently outperform strong baselines, emphasizing the value of integrating both features. Additionally, we demonstrate that these features enhance model robustness in scenarios with imbalanced labels and across different domains.

%% file: 8.limitations.tex
\section*{Limitations}
Our work has several limitations. First, the PDTB parser used in this study is far from perfect. Future research should focus on developing more powerful parsers to support discourse relation analysis for coherence modeling. For instance, it would be worthwhile to explore whether LLM-based approaches can produce better PDTB parsing results. Second, our experiments are limited to PDTB-style discourse relations. Extending the analysis to other frameworks, such as RST~\citep{RST}, could offer valuable insights. Finally, due to budget and computational constraints, we only experimented with Llama-8B (and only used Llama-70B in zero-shot setting). It would be interesting to evaluate our approach using other or larger language models, such as GPT-4.

%% file: 9.appendix.tex
\section{PDTB Parser}
\label{app:pdtb_parser}
\begin{table}[t]
\centering
\scalebox{0.69}{
\renewcommand{\arraystretch}{1.1}
\begin{tabular}{lc|lc}
\hline
Explicit           & Distribution & Implicit        & Distribution \\ \hline
Asynchronous       & 8.69\%        & Asynchronous    & 4.64\%        \\
Cause              & 7.87\%        & Cause           & 24.23\%        \\
Concession         & 19.94\%        & Cause+Belief    & 0.82\%        \\
Condition          & 5.99\%        & Concession      & 6.72\%        \\
Conjunction        & 36.55\%        & Condition       & 0.85\%        \\
Contrast           & 4.58\%        & Conjunction     & 20.84\%        \\
Disjunction        & 1.23\%        & Contrast        & 3.86\%        \\
Instantiation      & 1.30\%        & Equivalence     & 1.21\%        \\
Level-of-detail    & 1.01\%        & Instantiation   & 6.84\%        \\
Manner             & 1.23\%        & Level-of-detail & 14.60\%        \\
Negative-condition & 0.54\%        & Manner          & 0.74\%        \\
Purpose            & 1.63\%        & Purpose         & 3.31\%        \\
Similarity         & 0.42\%        & Substitution    & 1.34\%        \\
Substitution       & 0.96\%        & Synchronous     & 2.35\%        \\
Synchronous        & 8.07\%        & NoRel           & 8.18\%        \\ \hline
\end{tabular}}
\setlength{\abovecaptionskip}{10pt}
\setlength{\belowcaptionskip}{-6pt}
\caption{Explicit and Implicit relations used in this study and their distribution in the training corpus.}
\label{table-relations}
\end{table}

We use an updated version of {\tt discopy}~\citep{knaebel-2021-discopy} to parse discourse relations in documents. The first update involves replacing the PDTB 2.0~\citep{pdtb2} relation set with PDTB 3.0~\citep{pdtb3}. Specifically, we focus on identifying both explicit and implicit discourse relations between adjacent sentences. For explicit relations, we select 15 types that have sufficient training data~\citep{liu-etal-2023-hits,liu-etal-2024-causes}. For implicit relations, we include the 14 most frequent types, along with a ``NoRel'' label to account for cases where no relation is present—common in low-coherence texts. Table~\ref{table-relations} lists all the relations used in this study along with their distribution in PDTB 3.0.

The second update incorporates the model proposed by~\citet{liu-strube-2023-annotation} for recognizing implicit relations, due to its state-of-the-art performance. We implement the parser using RoBERTa and train it on PDTB 3.0, following the data split introduced by~\citet{ji-eisenstein-2015-one}. The parser achieves 89.61\% accuracy on the explicit test set and 67.80\% on the implicit test set of PDTB 3.0.

\begin{table}[t]
\centering
\scalebox{0.69}{
\renewcommand{\arraystretch}{1.1}
\begin{tabular}{l|l|lccc}
\hline
\multicolumn{2}{l|}{Dataset}                                           & Split & \#Doc & Avg \#Sent & Avg \#Word \\ \hline
\multicolumn{1}{l|}{\multirow{8}{*}{GCDC}}  & \multirow{2}{*}{Clinton} & Train & 1000  & 8.9        & 182.9      \\
\multicolumn{1}{l|}{}                       &                          & Test  & 200   & 8.8        & 186.0      \\
\multicolumn{1}{l|}{}                       & \multirow{2}{*}{Enron}   & Train & 1000  & 9.2        & 185.1      \\
\multicolumn{1}{l|}{}                       &                          & Test  & 200   & 9.3        & 191.1      \\
\multicolumn{1}{l|}{}                       & \multirow{2}{*}{Yahoo}   & Train & 1000  & 7.8        & 157.2      \\
\multicolumn{1}{l|}{}                       &                          & Test  & 200   & 7.8        & 162.7      \\
\multicolumn{1}{l|}{}                       & \multirow{2}{*}{Yelp}    & Train & 1000  & 10.4       & 178.2      \\
\multicolumn{1}{l|}{}                       &                          & Test  & 200   & 10.1       & 179.1      \\ \hline
CoheSentia & - & Total & 483 & 7.0 & 122.2 \\ \hline 
\multicolumn{1}{l|}{\multirow{8}{*}{TOEFL}} & Prompt 1                 & Total & 1656  & 13.7       & 339.1      \\
\multicolumn{1}{l|}{}                       & Prompt 2                 & Total & 1562  & 15.7       & 357.8      \\
\multicolumn{1}{l|}{}                       & Prompt 3                 & Total & 1396  & 14.7       & 343.5      \\
\multicolumn{1}{l|}{}                       & Prompt 4                 & Total & 1509  & 15.1       & 338.0      \\
\multicolumn{1}{l|}{}                       & Prompt 5                 & Total & 1648  & 15.2       & 358.4      \\
\multicolumn{1}{l|}{}                       & Prompt 6                 & Total & 960   & 15.3       & 358.3      \\
\multicolumn{1}{l|}{}                       & Prompt 7                 & Total & 1686  & 14.0       & 336.6      \\
\multicolumn{1}{l|}{}                       & Prompt 8                 & Total & 1683  & 14.7       & 340.9      \\ \hline
\end{tabular}}
\setlength{\abovecaptionskip}{10pt}
\setlength{\belowcaptionskip}{-6pt}
\caption{Statistics of datasets, where \#Doc, \#Sent, and \#Word mean the number of documents, sentences, and words, respectively.}
\label{table-stat}
\end{table}

\section{Experimental Settings}
\label{app:setting}
\subsection{Dataset}
The GCDC dataset includes texts from four domains: online forum posts from Yahoo, emails from the Enron corpus, emails from Hillary Clinton’s office, and online business reviews from Yelp. The CoheSentia datasets consists of stories generated by GPT-3. The TOEFL dataset comprises essays written in response to eight different prompts. Table~\ref{table-stat} presents statistics for these three corpora.

\subsection{Implementation}
\noindent \textbf{Fusion}. In the Fusion setting, we use a text encoder, such as RoBERTa or LLaMA, to obtain sentence representations. This is done by passing a sentence through the encoder, extracting token-level representations, and then averaging the representations of the tokens within the sentence. We experimented with both average pooling and [CLS] pooling methods. Our results show that average pooling consistently outperforms [CLS] pooling~\citep{liu-strube-2025-discourse}. For instance, on the TOEFL P1 dataset using a RoBERTa encoder, the accuracy of the TextOnly baseline and Our Method I with average pooling is 76.36 and 80.55, respectively, compared to 72.58 and 77.56 with [CLS] pooling. This improvement is likely because average pooling incorporates information from all tokens in the sentence, preserving more linguistic features. In contrast, [CLS] pooling relies solely on the [CLS] token's representation, which can result in the loss of important information. Similar results are observed for average pooling and [CLS] pooling in \citet{mosbach-etal-2020-interplay}. For entity and discourse relation elements in the flat structure, we convert them as vectors using GloVe embeddings~\citep{pennington-etal-2014-glove}. We use two layers of Fusion Transformers to jointly model sentences, entities, and discourse relations. Each layer consists of 8 attention heads and has a hidden size of 256. The model is trained using the AdamW optimizer with an initial learning rate of 1e-3, a batch size of 32, a dropout rate of 0.1, and a maximum of 20 training epochs.

\begin{figure*}[t]
\centering\includegraphics[scale=0.425,trim=0 0 0 0]{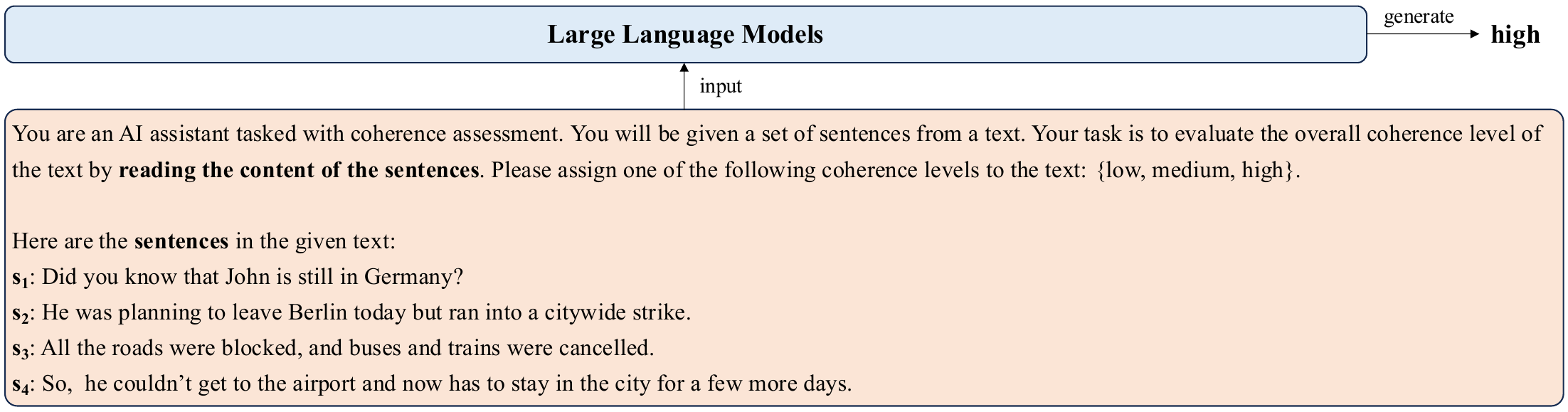}
\setlength{\abovecaptionskip}{-8pt}
\setlength{\belowcaptionskip}{1pt}
\caption{Illustration of TextOnly baseline in the Prompt setting. We instruct LLMs to consider only textual content for coherence assessment.}
\label{fig:plain_prompt}
\end{figure*}

\begin{table*}[t]
\centering
\scalebox{0.84}{
\renewcommand{\arraystretch}{1.2}
\begin{tabular}{lll|ccccc|c}
\hline
\multicolumn{3}{l|}{\multirow{2}{*}{Model}}                                                                        & \multicolumn{5}{c|}{\textbf{GCDC}}               & \multicolumn{1}{l}{\multirow{2}{*}{\textbf{CoheSentia}}} \\ \cline{4-8}
\multicolumn{3}{l|}{}                                                                                              & Clinton\hspace{0.6em} & Enron\hspace{0.8em} & Yahoo\hspace{0.8em} & Yelp\hspace{0.8em}  & Avg   & \multicolumn{1}{l}{}                            \\ \hline
\multicolumn{1}{l|}{\multirow{8}{*}{Fusion}} & \multicolumn{1}{l|}{\multirow{4}{*}{RoBERTa}}          & TextOnly   & 47.58\textsubscript{0.9} & 48.74\textsubscript{1.0} & 45.71\textsubscript{0.9} & 45.63\textsubscript{0.8} & 46.92 & 57.08\textsubscript{1.7}                                           \\
\multicolumn{1}{l|}{}                        & \multicolumn{1}{l|}{}                                  & TextEnty   & 52.38\textsubscript{1.2}   & 48.84\textsubscript{1.4} & 48.21\textsubscript{1.8} & 47.24\textsubscript{1.6} & 49.17 & 59.94\textsubscript{2.1}                                           \\
\multicolumn{1}{l|}{}                        & \multicolumn{1}{l|}{}                                  & TextRel    & 52.42\textsubscript{1.3}   & 51.04\textsubscript{1.5} & 48.56\textsubscript{1.7} & 47.35\textsubscript{1.8} & 49.84 & 60.35\textsubscript{1.9}                                           \\
\multicolumn{1}{l|}{}                        & \multicolumn{1}{l|}{}                                  & Our Method I & \textbf{54.49}\textsubscript{1.6} & 51.27\textsubscript{1.1} & 48.63\textsubscript{0.8} & 47.86\textsubscript{1.1} & 50.56 & 62.98\textsubscript{1.7}                                          \\ \cline{2-9} 
\multicolumn{1}{l|}{}                        & \multicolumn{1}{l|}{\multirow{4}{*}{Llama}}            & TextOnly   & 47.54\textsubscript{1.8}   & 48.73\textsubscript{1.6} & 44.38\textsubscript{1.0} & 46.09\textsubscript{1.4} & 46.68 & 59.95\textsubscript{1.6}                                           \\
\multicolumn{1}{l|}{}                        & \multicolumn{1}{l|}{}                                  & TextEnty   & 50.82\textsubscript{1.0}   & 50.98\textsubscript{1.1} & 47.74\textsubscript{0.8} & 47.29\textsubscript{1.4} & 49.20  & 62.52\textsubscript{2.0}                                           \\
\multicolumn{1}{l|}{}                        & \multicolumn{1}{l|}{}                                  & TextRel    & 49.73\textsubscript{1.7}   & 50.77\textsubscript{1.6} & 47.37\textsubscript{0.9} & \textbf{48.53}\textsubscript{0.6} & 49.10 & 63.67\textsubscript{2.1}                                           \\
\multicolumn{1}{l|}{}                        & \multicolumn{1}{l|}{}                                  & Our Method I & 53.78\textsubscript{1.3} & \textbf{52.37}\textsubscript{1.6} & \textbf{50.50}\textsubscript{1.3} & 47.59\textsubscript{1.3} & \textbf{51.06} & \textbf{65.25}\textsubscript{1.8}                                           \\ \hline \hline
\multicolumn{1}{l|}{\multirow{8}{*}{Prompt}} & \multicolumn{1}{l|}{\multirow{4}{*}{Llama zero-shot}}  & TextOnly   & 34.78\hspace{1.0em}   & 32.02\hspace{1.0em} & 32.39\hspace{1.0em} & 32.79\hspace{1.0em} & 33.88 & 40.06\hspace{1.0em}                                           \\
\multicolumn{1}{l|}{}                        & \multicolumn{1}{l|}{}                                  & TextEnty   & 40.24\hspace{1.0em}   & 34.71\hspace{1.0em} & 38.69\hspace{1.0em} & 36.56\hspace{1.0em} & 37.55 & 41.09\hspace{1.0em}                                           \\
\multicolumn{1}{l|}{}                        & \multicolumn{1}{l|}{}                                  & TextRel    & 41.43\hspace{1.0em}   & 36.37\hspace{1.0em} & 39.12\hspace{1.0em} & 36.56\hspace{1.0em} & 38.37 & 42.46\hspace{1.0em}                                           \\
\multicolumn{1}{l|}{}                        & \multicolumn{1}{l|}{}                                  & Our Method II & 41.74\hspace{1.0em}   & 34.40\hspace{1.0em} & 37.99\hspace{1.0em} & 40.14\hspace{1.0em} & 38.82 & 45.56\hspace{1.0em}                                           \\ \cline{2-9} 
\multicolumn{1}{l|}{}                        & \multicolumn{1}{l|}{\multirow{4}{*}{Llama fine-tuned}} & TextOnly   & 46.18\textsubscript{1.6}   & 44.83\textsubscript{1.1} & 46.41\textsubscript{1.4} & 38.21\textsubscript{1.3} & 43.90 & 57.46\textsubscript{1.7}                                          \\
\multicolumn{1}{l|}{}                        & \multicolumn{1}{l|}{}                                  & TextEnty   & 47.41\textsubscript{1.7}   & 45.37\textsubscript{1.5} & 46.69\textsubscript{1.6} & 39.18\textsubscript{1.2} & 44.66 & 58.36\textsubscript{1.8}                                          \\
\multicolumn{1}{l|}{}                        & \multicolumn{1}{l|}{}                                  & TextRel    & 46.91\textsubscript{1.5}   & 46.53\textsubscript{1.4} & 47.73\textsubscript{1.3} & 40.15\textsubscript{1.2} & 45.33 & 62.17\textsubscript{1.4}                                           \\
\multicolumn{1}{l|}{}                        & \multicolumn{1}{l|}{}                                  & Our Method II & 48.78\textsubscript{1.5}   & 49.46\textsubscript{1.3} & 48.23\textsubscript{1.3} & 41.00\textsubscript{0.9} & 46.87 & 63.65\textsubscript{1.5}                                          \\ \hline
\end{tabular}}
\setlength{\abovecaptionskip}{6pt}
\setlength{\belowcaptionskip}{-12pt}
\caption{Mean macro-F1 results (with std) on GCDC and CoheSentia.}
\label{table:gcdc_cohe_f1}
\end{table*}

\noindent \textbf{Prompt}. In the Prompt setting, the data is organized in the Alpaca format~\citep{dubois2023alpacafarm}. Our implementation is built on LlamaFactory~\citep{zheng-etal-2024-llamafactory}, a unified framework that incorporates a range of state-of-the-art efficient training methods for large language models (LLMs). In the zero-shot setting, we do not train the models; instead, we directly use LlamaFactory for evaluation. In the fine-tuned setting, we train using LoRA with a rank of 24, a LoRA alpha of 48, a dropout rate of 0.1, a learning rate of 5e-5, and a total of 3 training epochs.


\begin{table*}[t]
\centering
\Large
\scalebox{0.58}{
\renewcommand{\arraystretch}{1.3}
\begin{tabular}{lll|ccccccccc}
\hline
\multicolumn{3}{l|}{Model}                                                                                            & P1\hspace{0.9em}    & P2\hspace{0.9em}    & P3\hspace{0.9em}    & P4\hspace{0.9em}    & P5\hspace{0.9em}    & P6\hspace{0.9em}    & P7\hspace{0.9em}    & P8\hspace{0.9em}    & Avg   \\ \hline
\multicolumn{1}{l|}{\multirow{8}{*}{Fusion}} & \multicolumn{1}{l|}{\multirow{4}{*}{RoBERTa}}          & TextOnly      & 74.92\textsubscript{1.7} & 70.83\textsubscript{1.8} & 74.50\textsubscript{1.5} & 75.68\textsubscript{1.8} & 76.34\textsubscript{1.7} & 72.64\textsubscript{1.6} & 72.14\textsubscript{1.6} & 71.97\textsubscript{1.3} & 73.63 \\
\multicolumn{1}{l|}{}                        & \multicolumn{1}{l|}{}                                  & TextEnty      & 75.18\textsubscript{1.8} & 72.36\textsubscript{1.5} & 74.06\textsubscript{1.4} & 76.26\textsubscript{1.2} & 76.57\textsubscript{1.7} & 74.62\textsubscript{1.6} & 75.42\textsubscript{1.6} & 73.68\textsubscript{1.7} & 74.77\\
\multicolumn{1}{l|}{}                        & \multicolumn{1}{l|}{}                                  & TextRel       & 75.00\textsubscript{1.9} & 72.70\textsubscript{1.9} & 75.68\textsubscript{1.8} & 74.94\textsubscript{1.6} & 76.70\textsubscript{1.7} & 72.86\textsubscript{1.9} & 73.85\textsubscript{1.6} & 73.76\textsubscript{1.5} & 74.44 \\
\multicolumn{1}{l|}{}                        & \multicolumn{1}{l|}{}                                  & Our Method I  & \textbf{78.63}\textsubscript{0.9} & \textbf{75.33}\textsubscript{1.5} & \textbf{77.98}\textsubscript{0.6} & \textbf{77.11}\textsubscript{1.6} & 77.68\textsubscript{0.6} & \textbf{77.23}\textsubscript{1.3} & \textbf{75.90}\textsubscript{1.9} & \textbf{74.82}\textsubscript{1.5} & \textbf{76.84} \\ \cline{2-12} 
\multicolumn{1}{l|}{}                        & \multicolumn{1}{l|}{\multirow{4}{*}{Llama}}            & TextOnly      & 70.52\textsubscript{1.7} & 68.29\textsubscript{1.3} & 70.91\textsubscript{0.8} & 70.50\textsubscript{1.6} & 72.42\textsubscript{1.4} & 71.25\textsubscript{2.1} & 70.46\textsubscript{1.3} & 68.72\textsubscript{1.7} & 70.38 \\
\multicolumn{1}{l|}{}                        & \multicolumn{1}{l|}{}                                  & TextEnty      & 72.39\textsubscript{1.3} & 70.66\textsubscript{1.9} & 72.71\textsubscript{1.6} & 72.13\textsubscript{1.8} & 73.50\textsubscript{1.8} & 73.53\textsubscript{1.5} & 71.29\textsubscript{1.8} & 69.37\textsubscript{1.6} & 72.11 \\
\multicolumn{1}{l|}{}                        & \multicolumn{1}{l|}{}                                  & TextRel       & 72.30\textsubscript{1.5} & 71.59\textsubscript{1.3} & 72.98\textsubscript{0.6} & 72.12\textsubscript{1.8} & 72.36\textsubscript{1.8} & 72.50\textsubscript{1.8} & 71.41\textsubscript{1.5} & 70.57\textsubscript{1.4} & 71.98 \\
\multicolumn{1}{l|}{}                        & \multicolumn{1}{l|}{}                                  & Our Method I  & 74.30\textsubscript{1.4} & 73.97\textsubscript{2.0} & 74.48\textsubscript{1.1} & 73.76\textsubscript{1.4} & 75.48\textsubscript{2.4} & 75.96\textsubscript{1.6} & 73.82\textsubscript{1.8} & 72.54\textsubscript{2.0} & 74.16 \\ \hline \hline
\multicolumn{1}{l|}{\multirow{8}{*}{Prompt}} & \multicolumn{1}{l|}{\multirow{4}{*}{Llama zero-shot}}  & TextOnly      & 45.48\hspace{0.9em} & 50.80\hspace{0.9em} & 49.15\hspace{0.9em} & 47.17\hspace{0.9em} & 40.96\hspace{0.9em} & 48.88\hspace{0.9em} & 41.58\hspace{0.9em} & 47.17\hspace{0.9em} & 46.40 \\
\multicolumn{1}{l|}{}                        & \multicolumn{1}{l|}{}                                  & TextEnty      & 51.48\hspace{0.9em} & 48.48\hspace{0.9em} & 51.27\hspace{0.9em} & 49.16\hspace{0.9em} & 58.48\hspace{0.9em} & 52.95\hspace{0.9em} & 52.26\hspace{0.9em} & 50.48\hspace{0.9em} & 50.57 \\
\multicolumn{1}{l|}{}                        & \multicolumn{1}{l|}{}                                  & TextRel       & 50.37\hspace{0.9em} & 50.14\hspace{0.9em} & 51.09\hspace{0.9em} & 50.64\hspace{0.9em} & 51.28\hspace{0.9em} & 51.76\hspace{0.9em} & 52.56\hspace{0.9em} & 50.15\hspace{0.9em} & 51.00 \\
\multicolumn{1}{l|}{}                        & \multicolumn{1}{l|}{}                                  & Our Method II & 51.89\hspace{0.9em} & 50.70\hspace{0.9em} & 52.73\hspace{0.9em} & 50.87\hspace{0.9em} & 51.77\hspace{0.9em} & 53.06\hspace{0.9em} & 53.32\hspace{0.9em} & 51.35\hspace{0.9em} & 51.96 \\ \cline{2-12} 
\multicolumn{1}{l|}{}                        & \multicolumn{1}{l|}{\multirow{4}{*}{Llama fine-tuned}} & TextOnly      & 74.92\textsubscript{1.7} & 70.83\textsubscript{1.8} & 74.50\textsubscript{1.5} & 75.68\textsubscript{1.8} & 76.34\textsubscript{1.7} & 72.64\textsubscript{1.6} & 72.14\textsubscript{1.6} & 71.97\textsubscript{1.3} & 73.63 \\
\multicolumn{1}{l|}{}                        & \multicolumn{1}{l|}{}                                  & TextEnty      & 75.18\textsubscript{1.8} & 72.36\textsubscript{1.5} & 74.06\textsubscript{1.4} & 76.26\textsubscript{1.2} & 76.57\textsubscript{1.7} & 74.62\textsubscript{1.6} & 75.42\textsubscript{1.6} & 73.68\textsubscript{1.7} & 74.77 \\
\multicolumn{1}{l|}{}                        & \multicolumn{1}{l|}{}                                  & TextRel       & 75.00\textsubscript{1.9} & 72.70\textsubscript{1.9} & 75.68\textsubscript{1.8} & 74.94\textsubscript{1.6} & 76.70\textsubscript{1.7} & 72.86\textsubscript{1.9} & 73.85\textsubscript{1.6} & 73.76\textsubscript{1.5} & 74.44 \\
\multicolumn{1}{l|}{}                        & \multicolumn{1}{l|}{}                                  & Our Method II & 75.69\textsubscript{2.0} & 71.71\textsubscript{1.8} & 76.21\textsubscript{1.3} & 76.11\textsubscript{1.7} & \textbf{78.71}\textsubscript{1.7} & 74.82\textsubscript{1.5} & 73.82\textsubscript{2.0} & 74.48\textsubscript{1.7} & 75.19 \\ \hline
\end{tabular}}
\setlength{\abovecaptionskip}{6pt}
\setlength{\belowcaptionskip}{-2pt}
\caption{Mean macro-F1 results (with std) on TOEFL dataset.}
\label{table:toefl_f1}
\end{table*}

\begin{figure*}[t]
\centering\includegraphics[scale=0.425,trim=0 0 0 0]{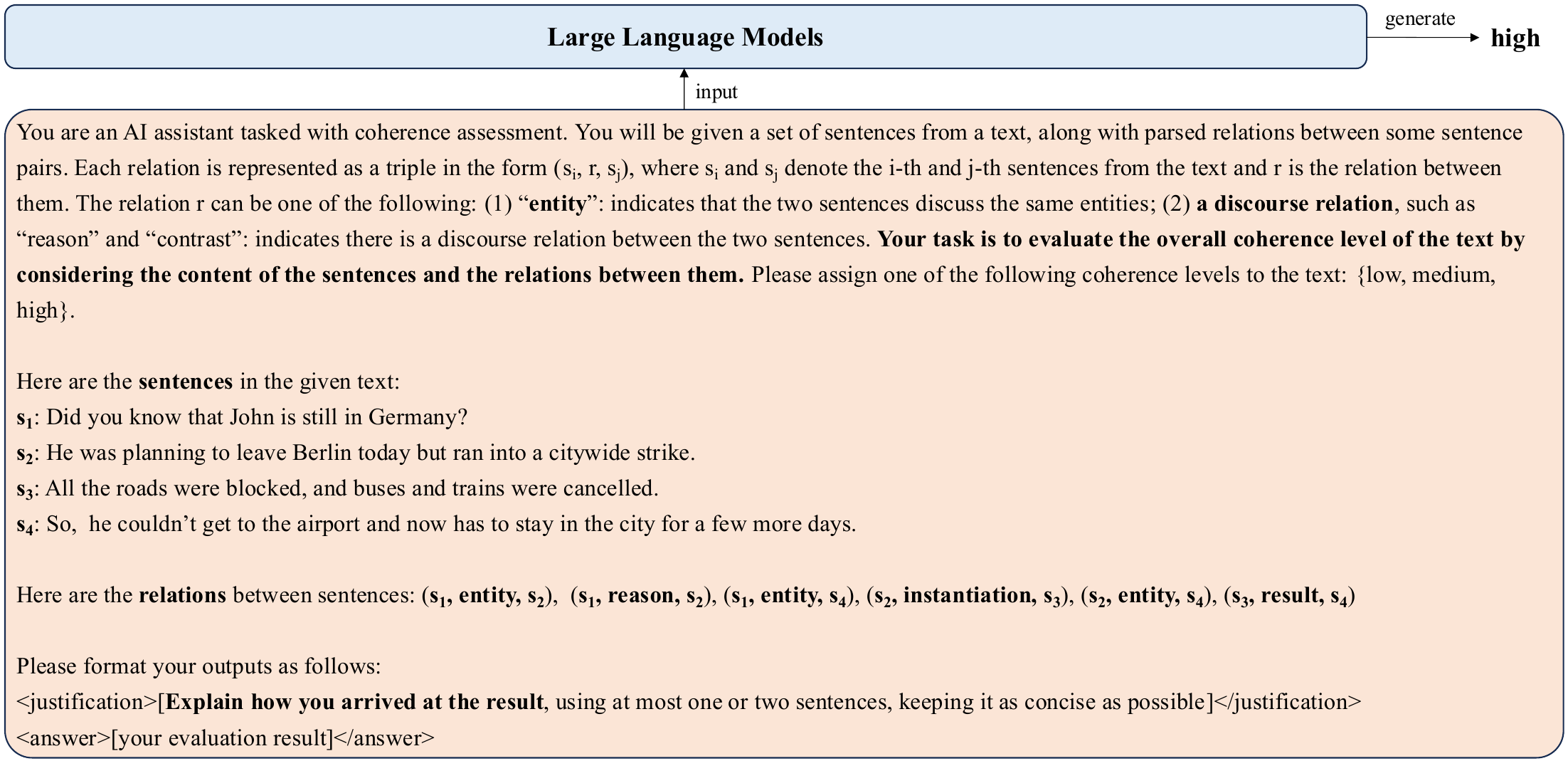}
\setlength{\abovecaptionskip}{-6pt}
\setlength{\belowcaptionskip}{-10pt}
\caption{Prompt with explanation.}
\label{fig:case_prompt}
\end{figure*}

\subsection{Baselines}
\noindent \textbf{TextOnly}. This baseline relies solely on textual content for coherence assessment. In the Fusion setting, we first use a text encoder to generate sentence representations, which are then passed through a sentence-level Transformer for feature extraction and finally fed into a Softmax layer for classification. Notably, no entities or discourse relations are used in this process. In the Prompt setting, we evaluate coherence by inputting only the text into large language models (LLMs). The prompt template used is shown in Figure \ref{fig:plain_prompt}.

\noindent \textbf{TextEnty}. This baseline is an ablated version of our approach. In the Fusion setting, we remove discourse relation elements from the flat structure, retaining only sentences and entities. In the Prompt setting, we include only triples connected by entity relations, such as ($\rm s_i$, entity, $\rm s_j$), in the prompt.

\noindent \textbf{TextRel}. This baseline is another ablated version of our approach. In the Fusion setting, we remove entity elements from the flat structure, retaining only sentences and discourse relations. In the Prompt setting, we include only triples connected by discourse relations, such as ($\rm s_i$, reason, $\rm s_j$), in the prompt.

\section{Macro-F1 Results}
\label{app:f1}
As noted in Section \ref{sec:analysis}, the labels in the GCDC, CoheSentia, and TOEFL corpora are imbalanced. While accuracy is commonly used as the evaluation metric for coherence assessment~\citep{lai-tetreault-2018-discourse,jeon-strube-2020-centering} and many other NLP tasks~\citep{fu-frank-2023-seti,fu-frank-2024-exploring,fu-frank-2024-compositional}, it does not account for the uneven label distribution~\citep{liu-etal-2019-encoding,liu-etal-2021-lexicon}. To address this, we also report model performance using Macro-F1, a standard metric for evaluating imbalanced datasets~\citep{macro-f1}. Tables \ref{table:gcdc_cohe_f1} and \ref{table:toefl_f1} present the results on the GCDC, CoheSentia, and TOEFL datasets. The trends in Macro-F1 scores closely mirror those observed in accuracy: incorporating entities and discourse relations improves performance, and combining both yields the best results.

\section{Prompt with Explanation}
\label{app:explanation}
In the case studies presented in Section \ref{sec:analysis}, we prompt LLaMA not only to evaluate the coherence level of a given text but also to provide a brief explanation for its judgment. This is done by modifying the instruction template used with LLaMA. Figure \ref{fig:case_prompt} shows the prompt used in these case studies for Our Method II. Similar prompts are used for TextOnly, TextEnty, and TextRel.

\begin{table*}[t!]
\centering
\scalebox{0.80}{
\renewcommand{\arraystretch}{1.2}
\begin{tabular}{lll|ccccc|c}
\hline
\multicolumn{3}{l|}{\multirow{2}{*}{Model}}                                                                                                                            & \multicolumn{5}{c|}{\textbf{GCDC}}               & \multirow{2}{*}{\textbf{CoheSentia}} \\ \cline{4-8}
\multicolumn{3}{l|}{}                                                                                                                                                  & Clinton & Enron & Yahoo & Yelp  & Avg   &                             \\ \hline
\multicolumn{1}{l|}{\multirow{4}{*}{Prompt}} & \multicolumn{1}{l|}{\multirow{4}{*}{\begin{tabular}[c]{@{}l@{}}Llama-3.3-70B\\ zero-shot\end{tabular}}} & TextOnly      & 56.50   & 51.00 & 43.50 & 47.50 & 49.63 & 55.07                       \\
\multicolumn{1}{l|}{}                        & \multicolumn{1}{l|}{}                                                                                   & TextEnty      & 57.50   & 51.50 & 45.50 & 52.00 & 51.63 & 56.11                       \\
\multicolumn{1}{l|}{}                        & \multicolumn{1}{l|}{}                                                                                   & TextRel       & 59.50   & 52.50 & 49.50 & 52.50 & 53.50 & 56.73                       \\
\multicolumn{1}{l|}{}                        & \multicolumn{1}{l|}{}                                                                                   & Our Method II & 60.00   & 53.50 & 52.50 & 53.00 & 54.75 & 57.56                       \\ \hline
\end{tabular}}
\setlength{\abovecaptionskip}{4pt}
\setlength{\belowcaptionskip}{-2pt}
\caption{Mean accuracy results of \textbf{Llama-3.3-70B} on GCDC and CoheSentia in the \textbf{zero-shot setting}.}
\label{table:gcdc_cohe_70b}
\end{table*}

\begin{table*}[t!]
\centering
\scalebox{0.80}{
\renewcommand{\arraystretch}{1.2}
\begin{tabular}{lll|ccccc|c}
\hline
\multicolumn{3}{l|}{\multirow{2}{*}{Model}}                                                                                                                            & \multicolumn{5}{c|}{\textbf{GCDC}}               & \multirow{2}{*}{\textbf{CoheSentia}} \\ \cline{4-8}
\multicolumn{3}{l|}{}                                                                                                                                                  & Clinton & Enron & Yahoo & Yelp  & Avg   &                             \\ \hline
\multicolumn{1}{l|}{\multirow{4}{*}{Prompt}} & \multicolumn{1}{l|}{\multirow{4}{*}{\begin{tabular}[c]{@{}l@{}}Llama-3.3-70B\\ zero-shot\end{tabular}}} & TextOnly      & 41.84   & 36.30 & 36.12 & 35.55 & 37.45 & 45.84                       \\
\multicolumn{1}{l|}{}                        & \multicolumn{1}{l|}{}                                                                                   & TextEnty      & 44.61   & 38.68 & 40.74 & 38.68 & 40.68 & 48.74                       \\
\multicolumn{1}{l|}{}                        & \multicolumn{1}{l|}{}                                                                                   & TextRel       & 45.69   & 41.42 & 42.83 & 39.74 & 42.42 & 48.46                       \\
\multicolumn{1}{l|}{}                        & \multicolumn{1}{l|}{}                                                                                   & Our Method II & 47.00   & 40.68 & 41.69 & 41.56 & 42.73 & 50.62                       \\ \hline
\end{tabular}}
\setlength{\abovecaptionskip}{4pt}
\setlength{\belowcaptionskip}{-2pt}
\caption{Mean macro-F1 results of \textbf{Llama-3.3-70B} on GCDC and CoheSentia in the \textbf{zero-shot setting}.}
\label{table:gcdc_cohe_70b_f1}
\end{table*}

\begin{table*}[t!]
\centering
\scalebox{0.78}{
\renewcommand{\arraystretch}{1.2}
\begin{tabular}{lll|ccccccccc}
\hline
\multicolumn{3}{l|}{Models}                                                                                                                                            & P1    & P2    & P3    & P4    & P5    & P6    & P7    & P8    & Avg   \\ \hline
\multicolumn{1}{l|}{\multirow{4}{*}{Prompt}} & \multicolumn{1}{l|}{\multirow{4}{*}{\begin{tabular}[c]{@{}l@{}}Llama-3.3-70B\\ zero-shot\end{tabular}}} & TextOnly      & 57.25 & 58.51 & 54.58 & 54.67 & 57.95 & 56.46 & 53.62 & 54.37 & 55.93 \\
\multicolumn{1}{l|}{}                        & \multicolumn{1}{l|}{}                                                                                   & TextEnty      & 60.51 & 58.26 & 56.30 & 58.05 & 58.25 & 60.42 & 60.26 & 56.80 & 58.61 \\
\multicolumn{1}{l|}{}                        & \multicolumn{1}{l|}{}                                                                                   & TextRel       & 61.05 & 59.35 & 56.88 & 58.45 & 59.83 & 60.21 & 61.33 & 56.51 & 59.20 \\
\multicolumn{1}{l|}{}                        & \multicolumn{1}{l|}{}                                                                                   & Our Method II & 62.56 & 60.24 & 59.74 & 59.91 & 61.35 & 62.19 & 61.80 & 58.23 & 60.75 \\ \hline
\end{tabular}}
\setlength{\abovecaptionskip}{4pt}
\setlength{\belowcaptionskip}{-2pt}
\caption{Mean accuracy results of \textbf{Llama-3.3-70B} on TOEFL dataset in the \textbf{zero-shot setting}.}
\label{table:toefl_70b}
\end{table*}

\begin{table*}[t!]
\centering
\scalebox{0.78}{
\renewcommand{\arraystretch}{1.2}
\begin{tabular}{lll|ccccccccc}
\hline
\multicolumn{3}{l|}{Models}                                                                                                                                            & P1    & P2    & P3    & P4    & P5    & P6    & P7    & P8    & Avg   \\ \hline
\multicolumn{1}{l|}{\multirow{4}{*}{Prompt}} & \multicolumn{1}{l|}{\multirow{4}{*}{\begin{tabular}[c]{@{}l@{}}Llama-3.3-70B\\ zero-shot\end{tabular}}} & TextOnly      & 48.28 & 52.18 & 51.06 & 49.55 & 48.29 & 52.45 & 48.43 & 51.70 & 50.24 \\
\multicolumn{1}{l|}{}                        & \multicolumn{1}{l|}{}                                                                                   & TextEnty      & 51.42 & 51.69 & 53.36 & 52.34 & 51.72 & 55.06 & 54.21 & 53.62 & 52.93 \\
\multicolumn{1}{l|}{}                        & \multicolumn{1}{l|}{}                                                                                   & TextRel       & 52.37 & 53.42 & 53.87 & 53.82 & 52.55 & 54.87 & 56.45 & 53.88 & 53.90 \\
\multicolumn{1}{l|}{}                        & \multicolumn{1}{l|}{}                                                                                   & Our Method II & 54.01 & 54.38 & 55.64 & 54.28 & 54.84 & 56.07 & 57.35 & 55.16 & 55.22 \\ \hline
\end{tabular}}
\setlength{\abovecaptionskip}{4pt}
\setlength{\belowcaptionskip}{-2pt}
\caption{Mean macro-F1 results of \textbf{Llama-3.3-70B} on TOEFL dataset in the \textbf{zero-shot setting}.}
\label{table:toefl_70b_f1}
\end{table*}

\section{Zero-shot results using LLama-3.3-70B}
\label{app:zero-shot}
Coherence assessment involves processing entire documents as input, which are typically quite lengthy (see Table \ref{table-stat}). As a result, training and inference require GPUs with substantial memory capacity. Due to hardware limitations, we employ LLaMA-3.1-8B as the language model for implementing Method II in Section \ref{sec:experiment}. Although we also experimented with the more advanced LLaMA-3.3-70B model, it caused out-of-memory errors during fine-tuning. However, our GPU is capable of running LLaMA-3.3-70B in a zero-shot setting for Method II. Accordingly, we report the zero-shot results (including Accuracy and Macro-F1) using LLaMA-3.3-70B in Tables \ref{table:gcdc_cohe_70b}, \ref{table:gcdc_cohe_70b_f1}, \ref{table:toefl_70b}, and \ref{table:toefl_70b_f1}. As shown, the results are consistent with those obtained using LLaMA-3.1-8B: incorporating entity and discourse relations improves the model’s performance in coherence assessment, and jointly modeling both types of information yields the best results.

%% file: main.bbl
\begin{thebibliography}{69}
\providecommand{\natexlab}[1]{#1}

\bibitem[{Abhishek et~al.(2021)Abhishek, Rawat, Gupta, and Varma}]{abhishek2021transformer}
Tushar Abhishek, Daksh Rawat, Manish Gupta, and Vasudeva Varma. 2021.
\newblock Transformer models for text coherence assessment.
\newblock \emph{arXiv preprint arXiv:2109.02176}.

\bibitem[{Barzilay and Lapata(2008)}]{barzilay-lapata-2008-modeling}
Regina Barzilay and Mirella Lapata. 2008.
\newblock \href {https://doi.org/10.1162/coli.2008.34.1.1} {Modeling local coherence: An entity-based approach}.
\newblock \emph{Computational Linguistics}, 34(1):1--34.

\bibitem[{Beyer et~al.(2021)Beyer, Lo{\'a}iciga, and Schlangen}]{beyer-etal-2021-incoherence}
Anne Beyer, Sharid Lo{\'a}iciga, and David Schlangen. 2021.
\newblock \href {https://doi.org/10.18653/v1/2021.naacl-main.328} {Is incoherence surprising? targeted evaluation of coherence prediction from language models}.
\newblock In \emph{Proceedings of the 2021 Conference of the North American Chapter of the Association for Computational Linguistics: Human Language Technologies}, pages 4164--4173, Online. Association for Computational Linguistics.

\bibitem[{Blanchard et~al.(2013)Blanchard, Tetreault, Higgins, Cahill, and Chodorow}]{toefl11}
Daniel Blanchard, Joel Tetreault, Derrick Higgins, Aoife Cahill, and Martin Chodorow. 2013.
\newblock \href {https://doi.org/10.1002/j.2333-8504.2013.tb02331.x} {Toefl11: A corpus of non-native english}.
\newblock \emph{ETS Research Report Series}, 2013(2):i--15.

\bibitem[{Burstein et~al.(2010)Burstein, Tetreault, and Andreyev}]{burstein-etal-2010-using}
Jill Burstein, Joel Tetreault, and Slava Andreyev. 2010.
\newblock \href {https://aclanthology.org/N10-1099/} {Using entity-based features to model coherence in student essays}.
\newblock In \emph{Human Language Technologies: The 2010 Annual Conference of the North {A}merican Chapter of the Association for Computational Linguistics}, pages 681--684, Los Angeles, California. Association for Computational Linguistics.

\bibitem[{Chaturvedi et~al.(2017)Chaturvedi, Peng, and Roth}]{chaturvedi-etal-2017-story}
Snigdha Chaturvedi, Haoruo Peng, and Dan Roth. 2017.
\newblock \href {https://doi.org/10.18653/v1/D17-1168} {Story comprehension for predicting what happens next}.
\newblock In \emph{Proceedings of the 2017 Conference on Empirical Methods in Natural Language Processing}, pages 1603--1614, Copenhagen, Denmark. Association for Computational Linguistics.

\bibitem[{Clark(1975)}]{clark-1975-bridging}
Herbert~H. Clark. 1975.
\newblock \href {https://aclanthology.org/T75-2034/} {Bridging}.
\newblock In \emph{Theoretical Issues in Natural Language Processing}.

\bibitem[{Dai et~al.(2019)Dai, Yang, Yang, Carbonell, Le, and Salakhutdinov}]{dai-etal-2019-transformer}
Zihang Dai, Zhilin Yang, Yiming Yang, Jaime Carbonell, Quoc Le, and Ruslan Salakhutdinov. 2019.
\newblock \href {https://doi.org/10.18653/v1/P19-1285} {Transformer-{XL}: Attentive language models beyond a fixed-length context}.
\newblock In \emph{Proceedings of the 57th Annual Meeting of the Association for Computational Linguistics}, pages 2978--2988, Florence, Italy. Association for Computational Linguistics.

\bibitem[{Dubois et~al.(2023)Dubois, Li, Taori, Zhang, Gulrajani, Ba, Guestrin, Liang, and Hashimoto}]{dubois2023alpacafarm}
Yann Dubois, Xuechen Li, Rohan Taori, Tianyi Zhang, Ishaan Gulrajani, Jimmy Ba, Carlos Guestrin, Percy Liang, and Tatsunori Hashimoto. 2023.
\newblock \href {https://openreview.net/forum?id=4hturzLcKX} {Alpacafarm: A simulation framework for methods that learn from human feedback}.
\newblock In \emph{Thirty-seventh Conference on Neural Information Processing Systems}.

\bibitem[{Elsner and Charniak(2011)}]{elsner-charniak-2011-extending}
Micha Elsner and Eugene Charniak. 2011.
\newblock \href {https://aclanthology.org/P11-2022/} {Extending the entity grid with entity-specific features}.
\newblock In \emph{Proceedings of the 49th Annual Meeting of the Association for Computational Linguistics: Human Language Technologies}, pages 125--129, Portland, Oregon, USA. Association for Computational Linguistics.

\bibitem[{Feng et~al.(2014)Feng, Lin, and Hirst}]{feng-etal-2014-impact}
Vanessa~Wei Feng, Ziheng Lin, and Graeme Hirst. 2014.
\newblock \href {https://aclanthology.org/C14-1089/} {The impact of deep hierarchical discourse structures in the evaluation of text coherence}.
\newblock In \emph{Proceedings of {COLING} 2014, the 25th International Conference on Computational Linguistics: Technical Papers}, pages 940--949, Dublin, Ireland. Dublin City University and Association for Computational Linguistics.

\bibitem[{Filippova and Strube(2007)}]{filippova-strube-2007-extending}
Katja Filippova and Michael Strube. 2007.
\newblock \href {https://aclanthology.org/W07-2321/} {Extending the entity-grid coherence model to semantically related entities}.
\newblock In \emph{Proceedings of the Eleventh {E}uropean Workshop on Natural Language Generation ({ENLG} 07)}, pages 139--142, Saarbr{\"u}cken, Germany. DFKI GmbH.

\bibitem[{Fu and Frank(2023)}]{fu-frank-2023-seti}
Xiyan Fu and Anette Frank. 2023.
\newblock \href {https://doi.org/10.18653/v1/2023.findings-acl.252} {{SETI}: Systematicity evaluation of textual inference}.
\newblock In \emph{Findings of the Association for Computational Linguistics: ACL 2023}, pages 4101--4114, Toronto, Canada. Association for Computational Linguistics.

\bibitem[{Fu and Frank(2024{\natexlab{a}})}]{fu-frank-2024-compositional}
Xiyan Fu and Anette Frank. 2024{\natexlab{a}}.
\newblock \href {https://doi.org/10.18653/v1/2024.starsem-1.31} {Compositional structured explanation generation with dynamic modularized reasoning}.
\newblock In \emph{Proceedings of the 13th Joint Conference on Lexical and Computational Semantics (*SEM 2024)}, pages 385--401, Mexico City, Mexico. Association for Computational Linguistics.

\bibitem[{Fu and Frank(2024{\natexlab{b}})}]{fu-frank-2024-exploring}
Xiyan Fu and Anette Frank. 2024{\natexlab{b}}.
\newblock \href {https://doi.org/10.1162/tacl_a_00680} {Exploring continual learning of compositional generalization in {NLI}}.
\newblock \emph{Transactions of the Association for Computational Linguistics}, 12:912--932.

\bibitem[{Grattafiori et~al.(2024)Grattafiori, Dubey, Jauhri, Pandey, Kadian, Al-Dahle, Letman, Mathur, Schelten, Vaughan, Yang, Fan, Goyal, Hartshorn, Yang, Mitra, Sravankumar, Korenev, Hinsvark, Rao, Zhang, Rodriguez, Gregerson, Spataru, Roziere, Biron, Tang, Chern, Caucheteux, Nayak, Bi, Marra, McConnell, Keller, Touret, Wu, Wong, Ferrer, Nikolaidis, Allonsius, Song, Pintz, Livshits, Wyatt, Esiobu, Choudhary, Mahajan, Garcia-Olano, Perino, Hupkes, Lakomkin, AlBadawy, Lobanova, Dinan, Smith, Radenovic, Guzmán, Zhang, Synnaeve, Lee, Anderson, Thattai, Nail, Mialon, Pang, Cucurell, Nguyen, Korevaar, Xu, Touvron, Zarov, Ibarra, Kloumann, Misra, Evtimov, Zhang, Copet, Lee, Geffert, Vranes, Park, Mahadeokar, Shah, van~der Linde, Billock, Hong, Lee, Fu, Chi, Huang, Liu, Wang, Yu, Bitton, Spisak, Park, Rocca, Johnstun, Saxe, Jia, Alwala, Prasad, Upasani, Plawiak, Li, Heafield, Stone, El-Arini, Iyer, Malik, Chiu, Bhalla, Lakhotia, Rantala-Yeary, van~der Maaten, Chen, Tan, Jenkins, Martin, Madaan, Malo, Blecher,
  Landzaat, de~Oliveira, Muzzi, Pasupuleti, Singh, Paluri, Kardas, Tsimpoukelli, Oldham, Rita, Pavlova, Kambadur, Lewis, Si, Singh, Hassan, Goyal, Torabi, Bashlykov, Bogoychev, Chatterji, Zhang, Duchenne, Çelebi, Alrassy, Zhang, Li, Vasic, Weng, Bhargava, Dubal, Krishnan, Koura, Xu, He, Dong, Srinivasan, Ganapathy, Calderer, Cabral, Stojnic, Raileanu, Maheswari, Girdhar, Patel, Sauvestre, Polidoro, Sumbaly, Taylor, Silva, Hou, Wang, Hosseini, Chennabasappa, Singh, Bell, Kim, Edunov, Nie, Narang, Raparthy, Shen, Wan, Bhosale, Zhang, Vandenhende, Batra, Whitman, Sootla, Collot, Gururangan, Borodinsky, Herman, Fowler, Sheasha, Georgiou, Scialom, Speckbacher, Mihaylov, Xiao, Karn, Goswami, Gupta, Ramanathan, Kerkez, Gonguet, Do, Vogeti, Albiero, Petrovic, Chu, Xiong, Fu, Meers, Martinet, Wang, Wang, Tan, Xia, Xie, Jia, Wang, Goldschlag, Gaur, Babaei, Wen, Song, Zhang, Li, Mao, Coudert, Yan, Chen, Papakipos, Singh, Srivastava, Jain, Kelsey, Shajnfeld, Gangidi, Victoria, Goldstand, Menon, Sharma, Boesenberg,
  Baevski, Feinstein, Kallet, Sangani, Teo, Yunus, Lupu, Alvarado, Caples, Gu, Ho, Poulton, Ryan, Ramchandani, Dong, Franco, Goyal, Saraf, Chowdhury, Gabriel, Bharambe, Eisenman, Yazdan, James, Maurer, Leonhardi, Huang, Loyd, Paola, Paranjape, Liu, Wu, Ni, Hancock, Wasti, Spence, Stojkovic, Gamido, Montalvo, Parker, Burton, Mejia, Liu, Wang, Kim, Zhou, Hu, Chu, Cai, Tindal, Feichtenhofer, Gao, Civin, Beaty, Kreymer, Li, Adkins, Xu, Testuggine, David, Parikh, Liskovich, Foss, Wang, Le, Holland, Dowling, Jamil, Montgomery, Presani, Hahn, Wood, Le, Brinkman, Arcaute, Dunbar, Smothers, Sun, Kreuk, Tian, Kokkinos, Ozgenel, Caggioni, Kanayet, Seide, Florez, Schwarz, Badeer, Swee, Halpern, Herman, Sizov, Guangyi, Zhang, Lakshminarayanan, Inan, Shojanazeri, Zou, Wang, Zha, Habeeb, Rudolph, Suk, Aspegren, Goldman, Zhan, Damlaj, Molybog, Tufanov, Leontiadis, Veliche, Gat, Weissman, Geboski, Kohli, Lam, Asher, Gaya, Marcus, Tang, Chan, Zhen, Reizenstein, Teboul, Zhong, Jin, Yang, Cummings, Carvill, Shepard, McPhie,
  Torres, Ginsburg, Wang, Wu, U, Saxena, Khandelwal, Zand, Matosich, Veeraraghavan, Michelena, Li, Jagadeesh, Huang, Chawla, Huang, Chen, Garg, A, Silva, Bell, Zhang, Guo, Yu, Moshkovich, Wehrstedt, Khabsa, Avalani, Bhatt, Mankus, Hasson, Lennie, Reso, Groshev, Naumov, Lathi, Keneally, Liu, Seltzer, Valko, Restrepo, Patel, Vyatskov, Samvelyan, Clark, Macey, Wang, Hermoso, Metanat, Rastegari, Bansal, Santhanam, Parks, White, Bawa, Singhal, Egebo, Usunier, Mehta, Laptev, Dong, Cheng, Chernoguz, Hart, Salpekar, Kalinli, Kent, Parekh, Saab, Balaji, Rittner, Bontrager, Roux, Dollar, Zvyagina, Ratanchandani, Yuvraj, Liang, Alao, Rodriguez, Ayub, Murthy, Nayani, Mitra, Parthasarathy, Li, Hogan, Battey, Wang, Howes, Rinott, Mehta, Siby, Bondu, Datta, Chugh, Hunt, Dhillon, Sidorov, Pan, Mahajan, Verma, Yamamoto, Ramaswamy, Lindsay, Lindsay, Feng, Lin, Zha, Patil, Shankar, Zhang, Zhang, Wang, Agarwal, Sajuyigbe, Chintala, Max, Chen, Kehoe, Satterfield, Govindaprasad, Gupta, Deng, Cho, Virk, Subramanian, Choudhury,
  Goldman, Remez, Glaser, Best, Koehler, Robinson, Li, Zhang, Matthews, Chou, Shaked, Vontimitta, Ajayi, Montanez, Mohan, Kumar, Mangla, Ionescu, Poenaru, Mihailescu, Ivanov, Li, Wang, Jiang, Bouaziz, Constable, Tang, Wu, Wang, Wu, Gao, Kleinman, Chen, Hu, Jia, Qi, Li, Zhang, Zhang, Adi, Nam, Yu, Wang, Zhao, Hao, Qian, Li, He, Rait, DeVito, Rosnbrick, Wen, Yang, Zhao, and Ma}]{grattafiori2024llama3herdmodels}
Aaron Grattafiori, Abhimanyu Dubey, Abhinav Jauhri, Abhinav Pandey, Abhishek Kadian, Ahmad Al-Dahle, Aiesha Letman, Akhil Mathur, Alan Schelten, Alex Vaughan, Amy Yang, Angela Fan, Anirudh Goyal, Anthony Hartshorn, Aobo Yang, Archi Mitra, Archie Sravankumar, Artem Korenev, Arthur Hinsvark, and 542 others. 2024.
\newblock \href {https://arxiv.org/abs/2407.21783} {The llama 3 herd of models}.
\newblock \emph{Preprint}, arXiv:2407.21783.

\bibitem[{Grosz et~al.(1995)Grosz, Joshi, and Weinstein}]{grosz-etal-1995-centering}
Barbara~J. Grosz, Aravind~K. Joshi, and Scott Weinstein. 1995.
\newblock \href {https://aclanthology.org/J95-2003/} {{C}entering: A framework for modeling the local coherence of discourse}.
\newblock \emph{Computational Linguistics}, 21(2):203--225.

\bibitem[{Guan et~al.(2023)Guan, Yang, Zhang, Hu, and Huang}]{guan2023generating}
Jian Guan, Zhenyu Yang, Rongsheng Zhang, Zhipeng Hu, and Minlie Huang. 2023.
\newblock \href {https://doi.org/10.1609/aaai.v37i11.26509} {Generating coherent narratives by learning dynamic and discrete entity states with a contrastive framework}.
\newblock In \emph{Proceedings of the AAAI Conference on Artificial Intelligence}, volume~37, pages 12836--12844.

\bibitem[{Guinaudeau and Strube(2013)}]{guinaudeau-strube-2013-graph}
Camille Guinaudeau and Michael Strube. 2013.
\newblock \href {https://aclanthology.org/P13-1010/} {Graph-based local coherence modeling}.
\newblock In \emph{Proceedings of the 51st Annual Meeting of the Association for Computational Linguistics (Volume 1: Long Papers)}, pages 93--103, Sofia, Bulgaria. Association for Computational Linguistics.

\bibitem[{Hou et~al.(2018)Hou, Markert, and Strube}]{hou-etal-2018-unrestricted}
Yufang Hou, Katja Markert, and Michael Strube. 2018.
\newblock \href {https://doi.org/10.1162/COLI_a_00315} {Unrestricted bridging resolution}.
\newblock \emph{Computational Linguistics}, 44(2):237--284.

\bibitem[{Jeon and Strube(2020)}]{jeon-strube-2020-centering}
Sungho Jeon and Michael Strube. 2020.
\newblock \href {https://doi.org/10.18653/v1/2020.emnlp-main.604} {Centering-based neural coherence modeling with hierarchical discourse segments}.
\newblock In \emph{Proceedings of the 2020 Conference on Empirical Methods in Natural Language Processing (EMNLP)}, pages 7458--7472, Online. Association for Computational Linguistics.

\bibitem[{Jeon and Strube(2022)}]{jeon-strube-2022-entity}
Sungho Jeon and Michael Strube. 2022.
\newblock \href {https://doi.org/10.18653/v1/2022.acl-long.537} {Entity-based neural local coherence modeling}.
\newblock In \emph{Proceedings of the 60th Annual Meeting of the Association for Computational Linguistics (Volume 1: Long Papers)}, pages 7787--7805, Dublin, Ireland. Association for Computational Linguistics.

\bibitem[{Ji and Eisenstein(2015)}]{ji-eisenstein-2015-one}
Yangfeng Ji and Jacob Eisenstein. 2015.
\newblock \href {https://doi.org/10.1162/tacl_a_00142} {One vector is not enough: Entity-augmented distributed semantics for discourse relations}.
\newblock \emph{Transactions of the Association for Computational Linguistics}, 3:329--344.

\bibitem[{Joty et~al.(2018)Joty, Mohiuddin, and Tien~Nguyen}]{joty-etal-2018-coherence}
Shafiq Joty, Muhammad~Tasnim Mohiuddin, and Dat Tien~Nguyen. 2018.
\newblock \href {https://doi.org/10.18653/v1/P18-1052} {Coherence modeling of asynchronous conversations: A neural entity grid approach}.
\newblock In \emph{Proceedings of the 56th Annual Meeting of the Association for Computational Linguistics (Volume 1: Long Papers)}, pages 558--568, Melbourne, Australia. Association for Computational Linguistics.

\bibitem[{Jurafsky and Martin(2025)}]{speechlanguage}
Daniel Jurafsky and James~H. Martin. 2025.
\newblock \href {https://web.stanford.edu/~jurafsky/slp3/} {\emph{Speech and Language Processing: An Introduction to Natural Language Processing, Computational Linguistics, and Speech Recognition with Language Models}}, 3rd edition.
\newblock Online manuscript released January 12, 2025.

\bibitem[{Kehler et~al.(2008)Kehler, Kertz, Rohde, and Elman}]{kehler08}
Andrew Kehler, Laura Kertz, Hannah Rohde, and Jeffrey~L. Elman. 2008.
\newblock \href {https://doi.org/10.1093/JOS/FFM018} {Coherence and coreference revisited}.
\newblock \emph{J. Semant.}, 25(1):1--44.

\bibitem[{Knaebel(2021)}]{knaebel-2021-discopy}
Ren{\'e} Knaebel. 2021.
\newblock \href {https://doi.org/10.18653/v1/2021.codi-main.12} {discopy: A neural system for shallow discourse parsing}.
\newblock In \emph{Proceedings of the 2nd Workshop on Computational Approaches to Discourse}, pages 128--133, Punta Cana, Dominican Republic and Online. Association for Computational Linguistics.

\bibitem[{Laban et~al.(2021)Laban, Dai, Bandarkar, and Hearst}]{laban-etal-2021-transformer}
Philippe Laban, Luke Dai, Lucas Bandarkar, and Marti~A. Hearst. 2021.
\newblock \href {https://doi.org/10.18653/v1/2021.acl-short.134} {Can transformer models measure coherence in text: Re-thinking the shuffle test}.
\newblock In \emph{Proceedings of the 59th Annual Meeting of the Association for Computational Linguistics and the 11th International Joint Conference on Natural Language Processing (Volume 2: Short Papers)}, pages 1058--1064, Online. Association for Computational Linguistics.

\bibitem[{Lai and Tetreault(2018)}]{lai-tetreault-2018-discourse}
Alice Lai and Joel Tetreault. 2018.
\newblock \href {https://doi.org/10.18653/v1/W18-5023} {Discourse coherence in the wild: A dataset, evaluation and methods}.
\newblock In \emph{Proceedings of the 19th Annual {SIG}dial Meeting on Discourse and Dialogue}, pages 214--223, Melbourne, Australia. Association for Computational Linguistics.

\bibitem[{Lapata(2003)}]{lapata-2003-probabilistic}
Mirella Lapata. 2003.
\newblock \href {https://doi.org/10.3115/1075096.1075165} {Probabilistic text structuring: Experiments with sentence ordering}.
\newblock In \emph{Proceedings of the 41st Annual Meeting of the Association for Computational Linguistics}, pages 545--552, Sapporo, Japan. Association for Computational Linguistics.

\bibitem[{Lapata and Barzilay(2005)}]{lapatacoh}
Mirella Lapata and Regina Barzilay. 2005.
\newblock Automatic evaluation of text coherence: models and representations.
\newblock In \emph{Proceedings of the 19th International Joint Conference on Artificial Intelligence}, IJCAI'05, page 1085–1090, San Francisco, CA, USA. Morgan Kaufmann Publishers Inc.

\bibitem[{Li et~al.(2023)Li, Gonzalez-Pizarro, Xing, Murray, and Carenini}]{li-etal-2023-diversity}
Raymond Li, Felipe Gonzalez-Pizarro, Linzi Xing, Gabriel Murray, and Giuseppe Carenini. 2023.
\newblock \href {https://doi.org/10.18653/v1/2023.acl-short.145} {Diversity-aware coherence loss for improving neural topic models}.
\newblock In \emph{Proceedings of the 61st Annual Meeting of the Association for Computational Linguistics (Volume 2: Short Papers)}, pages 1710--1722, Toronto, Canada. Association for Computational Linguistics.

\bibitem[{Lin et~al.(2011)Lin, Ng, and Kan}]{lin-etal-2011-automatically}
Ziheng Lin, Hwee~Tou Ng, and Min-Yen Kan. 2011.
\newblock \href {https://aclanthology.org/P11-1100/} {Automatically evaluating text coherence using discourse relations}.
\newblock In \emph{Proceedings of the 49th Annual Meeting of the Association for Computational Linguistics: Human Language Technologies}, pages 997--1006, Portland, Oregon, USA. Association for Computational Linguistics.

\bibitem[{Liu et~al.(2023{\natexlab{a}})Liu, Fan, and Strube}]{liu-etal-2023-hits}
Wei Liu, Yi~Fan, and Michael Strube. 2023{\natexlab{a}}.
\newblock \href {https://doi.org/10.18653/v1/2023.disrpt-1.4} {{HITS} at {DISRPT} 2023: Discourse segmentation, connective detection, and relation classification}.
\newblock In \emph{Proceedings of the 3rd Shared Task on Discourse Relation Parsing and Treebanking (DISRPT 2023)}, pages 43--49, Toronto, Canada. The Association for Computational Linguistics.

\bibitem[{Liu et~al.(2023{\natexlab{b}})Liu, Fu, and Strube}]{liu-etal-2023-modeling}
Wei Liu, Xiyan Fu, and Michael Strube. 2023{\natexlab{b}}.
\newblock \href {https://doi.org/10.18653/v1/2023.acl-long.431} {Modeling structural similarities between documents for coherence assessment with graph convolutional networks}.
\newblock In \emph{Proceedings of the 61st Annual Meeting of the Association for Computational Linguistics (Volume 1: Long Papers)}, pages 7792--7808, Toronto, Canada. Association for Computational Linguistics.

\bibitem[{Liu et~al.(2021)Liu, Fu, Zhang, and Xiao}]{liu-etal-2021-lexicon}
Wei Liu, Xiyan Fu, Yue Zhang, and Wenming Xiao. 2021.
\newblock \href {https://doi.org/10.18653/v1/2021.acl-long.454} {Lexicon enhanced {C}hinese sequence labeling using {BERT} adapter}.
\newblock In \emph{Proceedings of the 59th Annual Meeting of the Association for Computational Linguistics and the 11th International Joint Conference on Natural Language Processing (Volume 1: Long Papers)}, pages 5847--5858, Online. Association for Computational Linguistics.

\bibitem[{Liu and Strube(2023)}]{liu-strube-2023-annotation}
Wei Liu and Michael Strube. 2023.
\newblock \href {https://doi.org/10.18653/v1/2023.acl-long.874} {Annotation-inspired implicit discourse relation classification with auxiliary discourse connective generation}.
\newblock In \emph{Proceedings of the 61st Annual Meeting of the Association for Computational Linguistics (Volume 1: Long Papers)}, pages 15696--15712, Toronto, Canada. Association for Computational Linguistics.

\bibitem[{Liu and Strube(2025)}]{liu-strube-2025-discourse}
Wei Liu and Michael Strube. 2025.
\newblock \href {https://doi.org/10.18653/v1/2025.acl-long.236} {Discourse relation-enhanced neural coherence modeling}.
\newblock In \emph{Proceedings of the 63rd Annual Meeting of the Association for Computational Linguistics (Volume 1: Long Papers)}, pages 4748--4762, Vienna, Austria. Association for Computational Linguistics.

\bibitem[{Liu et~al.(2024)Liu, Wan, and Strube}]{liu-etal-2024-causes}
Wei Liu, Stephen Wan, and Michael Strube. 2024.
\newblock \href {https://doi.org/10.18653/v1/2024.naacl-long.150} {What causes the failure of explicit to implicit discourse relation recognition?}
\newblock In \emph{Proceedings of the 2024 Conference of the North American Chapter of the Association for Computational Linguistics: Human Language Technologies (Volume 1: Long Papers)}, pages 2738--2753, Mexico City, Mexico. Association for Computational Linguistics.

\bibitem[{Liu et~al.(2019{\natexlab{a}})Liu, Xu, Xu, Song, and Zu}]{liu-etal-2019-encoding}
Wei Liu, Tongge Xu, Qinghua Xu, Jiayu Song, and Yueran Zu. 2019{\natexlab{a}}.
\newblock \href {https://doi.org/10.18653/v1/N19-1247} {An encoding strategy based word-character {LSTM} for {C}hinese {NER}}.
\newblock In \emph{Proceedings of the 2019 Conference of the North {A}merican Chapter of the Association for Computational Linguistics: Human Language Technologies, Volume 1 (Long and Short Papers)}, pages 2379--2389, Minneapolis, Minnesota. Association for Computational Linguistics.

\bibitem[{Liu et~al.(2019{\natexlab{b}})Liu, Ott, Goyal, Du, Joshi, Chen, Levy, Lewis, Zettlemoyer, and Stoyanov}]{roberta}
Yinhan Liu, Myle Ott, Naman Goyal, Jingfei Du, Mandar Joshi, Danqi Chen, Omer Levy, Mike Lewis, Luke Zettlemoyer, and Veselin Stoyanov. 2019{\natexlab{b}}.
\newblock \href {https://arxiv.org/abs/1907.11692} {{RoBERTa}: {A} robustly optimized {BERT} {P}retraining {A}pproach}.
\newblock \emph{CoRR}, abs/1907.11692.

\bibitem[{Maimon and Tsarfaty(2023)}]{maimon-tsarfaty-2023-cohesentia}
Aviya Maimon and Reut Tsarfaty. 2023.
\newblock \href {https://doi.org/10.18653/v1/2023.emnlp-main.324} {{COHESENTIA}: A novel benchmark of incremental versus holistic assessment of coherence in generated texts}.
\newblock In \emph{Proceedings of the 2023 Conference on Empirical Methods in Natural Language Processing}, pages 5328--5343, Singapore. Association for Computational Linguistics.

\bibitem[{Mann and Thompson(1988)}]{RST}
William~C. Mann and Sandra~A. Thompson. 1988.
\newblock \href {https://doi.org/doi:10.1515/text.1.1988.8.3.243} {Rhetorical structure theory: Toward a functional theory of text organization}.
\newblock \emph{Text - Interdisciplinary Journal for the Study of Discourse}, 8(3):243--281.

\bibitem[{Mansour et~al.(2024)Mansour, Albatarni, Eltanbouly, and Elsayed}]{mansour-etal-2024-large}
Watheq~Ahmad Mansour, Salam Albatarni, Sohaila Eltanbouly, and Tamer Elsayed. 2024.
\newblock \href {https://aclanthology.org/2024.lrec-main.247/} {Can large language models automatically score proficiency of written essays?}
\newblock In \emph{Proceedings of the 2024 Joint International Conference on Computational Linguistics, Language Resources and Evaluation (LREC-COLING 2024)}, pages 2777--2786, Torino, Italia. ELRA and ICCL.

\bibitem[{Mendonca et~al.(2024)Mendonca, Trancoso, and Lavie}]{mendonca-etal-2024-ecoh}
John Mendonca, Isabel Trancoso, and Alon Lavie. 2024.
\newblock \href {https://doi.org/10.18653/v1/2024.sigdial-1.44} {{EC}oh: Turn-level coherence evaluation for multilingual dialogues}.
\newblock In \emph{Proceedings of the 25th Annual Meeting of the Special Interest Group on Discourse and Dialogue}, pages 516--532, Kyoto, Japan. Association for Computational Linguistics.

\bibitem[{Mesgar and Strube(2015)}]{mesgar-strube-2015-graph}
Mohsen Mesgar and Michael Strube. 2015.
\newblock \href {https://doi.org/10.18653/v1/S15-1036} {Graph-based coherence modeling for assessing readability}.
\newblock In \emph{Proceedings of the Fourth Joint Conference on Lexical and Computational Semantics}, pages 309--318, Denver, Colorado. Association for Computational Linguistics.

\bibitem[{Mesgar and Strube(2016)}]{mesgar-strube-2016-lexical}
Mohsen Mesgar and Michael Strube. 2016.
\newblock \href {https://doi.org/10.18653/v1/N16-1167} {Lexical coherence graph modeling using word embeddings}.
\newblock In \emph{Proceedings of the 2016 Conference of the North {A}merican Chapter of the Association for Computational Linguistics: Human Language Technologies}, pages 1414--1423, San Diego, California. Association for Computational Linguistics.

\bibitem[{Mosbach et~al.(2020)Mosbach, Khokhlova, Hedderich, and Klakow}]{mosbach-etal-2020-interplay}
Marius Mosbach, Anna Khokhlova, Michael~A. Hedderich, and Dietrich Klakow. 2020.
\newblock \href {https://doi.org/10.18653/v1/2020.blackboxnlp-1.7} {On the interplay between fine-tuning and sentence-level probing for linguistic knowledge in pre-trained transformers}.
\newblock In \emph{Proceedings of the Third BlackboxNLP Workshop on Analyzing and Interpreting Neural Networks for NLP}, pages 68--82, Online. Association for Computational Linguistics.

\bibitem[{Naismith et~al.(2023)Naismith, Mulcaire, and Burstein}]{naismith-etal-2023-automated}
Ben Naismith, Phoebe Mulcaire, and Jill Burstein. 2023.
\newblock \href {https://doi.org/10.18653/v1/2023.bea-1.32} {Automated evaluation of written discourse coherence using {GPT}-4}.
\newblock In \emph{Proceedings of the 18th Workshop on Innovative Use of NLP for Building Educational Applications (BEA 2023)}, pages 394--403, Toronto, Canada. Association for Computational Linguistics.

\bibitem[{Opitz and Burst(2019)}]{macro-f1}
Juri Opitz and Sebastian Burst. 2019.
\newblock \href {https://arxiv.org/abs/1911.03347} {Macro {F1} and macro {F1}}.
\newblock \emph{CoRR}, abs/1911.03347.

\bibitem[{Parmar et~al.(2024)Parmar, Deilamsalehy, Dernoncourt, Yoon, Rossi, and Bui}]{parmar-etal-2024-towards}
Mihir Parmar, Hanieh Deilamsalehy, Franck Dernoncourt, Seunghyun Yoon, Ryan~A. Rossi, and Trung Bui. 2024.
\newblock \href {https://doi.org/10.18653/v1/2024.emnlp-main.1106} {Towards enhancing coherence in extractive summarization: Dataset and experiments with {LLM}s}.
\newblock In \emph{Proceedings of the 2024 Conference on Empirical Methods in Natural Language Processing}, pages 19810--19820, Miami, Florida, USA. Association for Computational Linguistics.

\bibitem[{Pennington et~al.(2014)Pennington, Socher, and Manning}]{pennington-etal-2014-glove}
Jeffrey Pennington, Richard Socher, and Christopher Manning. 2014.
\newblock \href {https://doi.org/10.3115/v1/D14-1162} {{G}lo{V}e: Global vectors for word representation}.
\newblock In \emph{Proceedings of the 2014 Conference on Empirical Methods in Natural Language Processing ({EMNLP})}, pages 1532--1543, Doha, Qatar. Association for Computational Linguistics.

\bibitem[{Prasad et~al.(2008)Prasad, Dinesh, Lee, Miltsakaki, Robaldo, Joshi, and Webber}]{pdtb2}
Rashmi Prasad, Nikhil Dinesh, Alan Lee, Eleni Miltsakaki, Livio Robaldo, Aravind Joshi, and Bonnie Webber. 2008.
\newblock \href {https://aclanthology.org/L08-1093/} {The {P}enn {D}iscourse {T}ree{B}ank 2.0.}
\newblock In \emph{Proceedings of the Sixth International Conference on Language Resources and Evaluation ({LREC}`08)}, Marrakech, Morocco. European Language Resources Association (ELRA).

\bibitem[{Prince(1981)}]{Prince_Taxonomy81}
Ellen~F. Prince. 1981.
\newblock Toward a taxonomy of given-new information.
\newblock In Peter Cole, editor, \emph{Radical Pragmatics}, pages 223--255. Academic Press, New York.

\bibitem[{Qi et~al.(2020)Qi, Zhang, Zhang, Bolton, and Manning}]{qi-etal-2020-stanza}
Peng Qi, Yuhao Zhang, Yuhui Zhang, Jason Bolton, and Christopher~D. Manning. 2020.
\newblock \href {https://doi.org/10.18653/v1/2020.acl-demos.14} {{S}tanza: A python natural language processing toolkit for many human languages}.
\newblock In \emph{Proceedings of the 58th Annual Meeting of the Association for Computational Linguistics: System Demonstrations}, pages 101--108, Online. Association for Computational Linguistics.

\bibitem[{Reinhart(1980)}]{coherencecondition}
Tanya Reinhart. 1980.
\newblock \href {http://www.jstor.org/stable/1771893} {Conditions for text coherence}.
\newblock \emph{Poetics Today}, 1(4):161--180.

\bibitem[{Rohde et~al.(2018)Rohde, Johnson, Schneider, and Webber}]{rohde-etal-2018-discourse}
Hannah Rohde, Alexander Johnson, Nathan Schneider, and Bonnie Webber. 2018.
\newblock \href {https://doi.org/10.18653/v1/P18-1210} {Discourse coherence: Concurrent explicit and implicit relations}.
\newblock In \emph{Proceedings of the 56th Annual Meeting of the Association for Computational Linguistics (Volume 1: Long Papers)}, pages 2257--2267, Melbourne, Australia. Association for Computational Linguistics.

\bibitem[{Sia and Duh(2023)}]{sia-duh-2023-context}
Suzanna Sia and Kevin Duh. 2023.
\newblock \href {https://aclanthology.org/2023.mtsummit-research.15/} {In-context learning as maintaining coherency: A study of on-the-fly machine translation using large language models}.
\newblock In \emph{Proceedings of Machine Translation Summit XIX, Vol. 1: Research Track}, pages 173--185, Macau SAR, China. Asia-Pacific Association for Machine Translation.

\bibitem[{Taghipour and Ng(2016)}]{taghipour-ng-2016-neural}
Kaveh Taghipour and Hwee~Tou Ng. 2016.
\newblock \href {https://doi.org/10.18653/v1/D16-1193} {A neural approach to automated essay scoring}.
\newblock In \emph{Proceedings of the 2016 Conference on Empirical Methods in Natural Language Processing}, pages 1882--1891, Austin, Texas. Association for Computational Linguistics.

\bibitem[{Tien~Nguyen and Joty(2017)}]{tien-nguyen-joty-2017-neural}
Dat Tien~Nguyen and Shafiq Joty. 2017.
\newblock \href {https://doi.org/10.18653/v1/P17-1121} {A neural local coherence model}.
\newblock In \emph{Proceedings of the 55th Annual Meeting of the Association for Computational Linguistics (Volume 1: Long Papers)}, pages 1320--1330, Vancouver, Canada. Association for Computational Linguistics.

\bibitem[{Vaswani et~al.(2017)Vaswani, Shazeer, Parmar, Uszkoreit, Jones, Gomez, Kaiser, and Polosukhin}]{transformer}
Ashish Vaswani, Noam Shazeer, Niki Parmar, Jakob Uszkoreit, Llion Jones, Aidan~N Gomez, \L~ukasz Kaiser, and Illia Polosukhin. 2017.
\newblock \href {https://proceedings.neurips.cc/paper_files/paper/2017/file/3f5ee243547dee91fbd053c1c4a845aa-Paper.pdf} {Attention is all you need}.
\newblock In \emph{Advances in Neural Information Processing Systems}, volume~30. Curran Associates, Inc.

\bibitem[{Wang et~al.(2019)Wang, Gyawali, Bruno, Molloy, Evanini, and Zechner}]{wang-etal-2019-using}
Xinhao Wang, Binod Gyawali, James~V. Bruno, Hillary~R. Molloy, Keelan Evanini, and Klaus Zechner. 2019.
\newblock \href {https://doi.org/10.18653/v1/W19-2719} {Using {R}hetorical {S}tructure {T}heory to assess discourse coherence for non-native spontaneous speech}.
\newblock In \emph{Proceedings of the Workshop on Discourse Relation Parsing and Treebanking 2019}, pages 153--162, Minneapolis, MN. Association for Computational Linguistics.

\bibitem[{Webber et~al.(2019)Webber, Prasad, Lee, and Joshi}]{pdtb3}
Bonnie Webber, Rashmi Prasad, Alan Lee, and Aravind Joshi. 2019.
\newblock The {P}enn {D}iscourse {T}ree{B}ank 3.0 annotation manual.
\newblock \emph{Philadelphia, University of Pennsylvania}, 35:108.

\bibitem[{Wu et~al.(2023)Wu, Shen, Lan, Mao, Bai, and Wu}]{wu-etal-2023-multi-task}
Hongyi Wu, Xinshu Shen, Man Lan, Shaoguang Mao, Xiaopeng Bai, and Yuanbin Wu. 2023.
\newblock \href {https://doi.org/10.18653/v1/2023.emnlp-main.412} {A multi-task dataset for assessing discourse coherence in {C}hinese essays: Structure, theme, and logic analysis}.
\newblock In \emph{Proceedings of the 2023 Conference on Empirical Methods in Natural Language Processing}, pages 6673--6688, Singapore. Association for Computational Linguistics.

\bibitem[{Wu et~al.(2021)Wu, Pan, Chen, Long, Zhang, and Yu}]{SurveyGNN}
Zonghan Wu, Shirui Pan, Fengwen Chen, Guodong Long, Chengqi Zhang, and Philip~S. Yu. 2021.
\newblock \href {https://doi.org/10.1109/TNNLS.2020.2978386} {A comprehensive survey on graph neural networks}.
\newblock \emph{IEEE Transactions on Neural Networks and Learning Systems}, 32(1):4--24.

\bibitem[{Xiao et~al.(2021)Xiao, Huber, and Carenini}]{xiao-etal-2021-predicting}
Wen Xiao, Patrick Huber, and Giuseppe Carenini. 2021.
\newblock \href {https://doi.org/10.18653/v1/2021.naacl-main.326} {Predicting discourse trees from transformer-based neural summarizers}.
\newblock In \emph{Proceedings of the 2021 Conference of the North American Chapter of the Association for Computational Linguistics: Human Language Technologies}, pages 4139--4152, Online. Association for Computational Linguistics.

\bibitem[{Ye et~al.(2024)Ye, Zhang, Wang, Xu, and Zhang}]{ye-etal-2024-language}
Ruosong Ye, Caiqi Zhang, Runhui Wang, Shuyuan Xu, and Yongfeng Zhang. 2024.
\newblock \href {https://aclanthology.org/2024.findings-eacl.132/} {Language is all a graph needs}.
\newblock In \emph{Findings of the Association for Computational Linguistics: EACL 2024}, pages 1955--1973, St. Julian{'}s, Malta. Association for Computational Linguistics.

\bibitem[{Zhao et~al.(2023)Zhao, Strube, and Eger}]{zhao-etal-2023-discoscore}
Wei Zhao, Michael Strube, and Steffen Eger. 2023.
\newblock \href {https://doi.org/10.18653/v1/2023.eacl-main.278} {{D}isco{S}core: Evaluating text generation with {BERT} and discourse coherence}.
\newblock In \emph{Proceedings of the 17th Conference of the European Chapter of the Association for Computational Linguistics}, pages 3865--3883, Dubrovnik, Croatia. Association for Computational Linguistics.

\bibitem[{Zheng et~al.(2024)Zheng, Zhang, Zhang, Ye, and Luo}]{zheng-etal-2024-llamafactory}
Yaowei Zheng, Richong Zhang, Junhao Zhang, Yanhan Ye, and Zheyan Luo. 2024.
\newblock \href {https://doi.org/10.18653/v1/2024.acl-demos.38} {{L}lama{F}actory: Unified efficient fine-tuning of 100+ language models}.
\newblock In \emph{Proceedings of the 62nd Annual Meeting of the Association for Computational Linguistics (Volume 3: System Demonstrations)}, pages 400--410, Bangkok, Thailand. Association for Computational Linguistics.

\end{thebibliography}
